\documentclass[11pt]{article}
\usepackage{acl}
\usepackage{times}
\usepackage{placeins}
\usepackage{latexsym}
\usepackage[T1]{fontenc}
\usepackage[utf8]{inputenc}
\usepackage{microtype}
\usepackage{inconsolata}
\usepackage{graphicx}
\usepackage{booktabs}
\usepackage{multirow}
\usepackage{amsmath}
\usepackage{amssymb}
\usepackage{graphicx}
\usepackage{algorithm}
\usepackage{algpseudocode}
\usepackage[table]{xcolor}
\usepackage{colortbl}
\definecolor{g1}{HTML}{e8f5e3}
\definecolor{g2}{HTML}{c2e4b5}
\definecolor{g3}{HTML}{88c96a}
\definecolor{g4}{HTML}{4aa822}
\definecolor{g5}{HTML}{2a7a04}
\usepackage{pifont}  % for \ding{51} and \ding{55}
\newcommand\blfootnote[1]{%
  \begingroup\renewcommand\thefootnote{}\footnote{#1}%
  \addtocounter{footnote}{-1}\endgroup}
\newcommand{\ck}{\checkmark}
\title{What Makes a Medical Checker Trainable?
Diagnosing Signal Collapse and Reward Hacking in
Checker-Guided RAG for Biomedical QA}

\author{
  Yuelyu Ji \quad Min Gu Kwak \quad Hang Zhang \quad Xizhi Wu \quad Chenyu Li \quad Yanshan Wang \\
  University of Pittsburgh, Pittsburgh, PA, USA \\
  \texttt{yueluji@gmail.com}
}

\begin{document}
\maketitle
\blfootnote{Code, prompts, and diagnostics available at: 
\url{https://anonymous.4open.science/r/medchecker-verl-7FC1/}}
% ============================================================
%  ABSTRACT  (~200 words)
% ============================================================
% \begin{abstract}
% Medical RAG needs evidence-grounded claims, so plugging a claim-level NLI checker into retrieval-augmented RL is intuitive. But the verifier's \emph{output distribution}, not its accuracy, decides whether it provides trainable gradient. We compare four NLI checker back-ends as process rewards inside a GRPO-trained medical RAG agent on Qwen2.5-7B (with Qwen3-4B and Llama-3.1-8B replications) across four held-out medical QA benchmarks (1{,}479 samples). Three diagnostic findings emerge. \textbf{(i) Signal collapse is log-prob-specific}: LLM log-probability scoring labels $>$97\% of claims neutral while MedNLI-Cls scores the same pairs non-degenerately at 54\% support. \textbf{(ii) Moderate signal beats strong signal on answer quality}: the 54\%-support classifier reaches
% BERTScore 0.600 ($+12$\% over zero-shot, no GPT
% dependency) while a 86\%-support GPT checker reaches
% 0.591, because strong signal triggers a three-step
% reward hacking cascade: ultra-short answers, search
% avoidance, language collapse.
% \textbf{(iii) Signal strength is policy-dependent}:
% the same checker registers as moderate on Qwen2.5-7B
% but strong on Qwen3-4B without triggering the cascade
% end-state.
% \end{abstract}% 
\begin{abstract}
Medical RAG needs evidence-grounded claims, so 
plugging a claim-level NLI checker into 
retrieval-augmented RL is intuitive. \textbf{We find 
that the checker's \emph{output distribution} during 
training, not its held-out accuracy, decides whether 
it provides trainable gradient.} We compare four NLI 
checker back-ends as process rewards inside a 
GRPO-trained medical RAG agent (Qwen2.5-7B, 
replicated on Qwen3-4B and Llama-3.1-8B) across four 
held-out medical QA benchmarks. Three diagnostic 
findings emerge. \textbf{(i)} Signal collapse is 
log-prob-specific: LLM log-probability scoring labels 
over 97\% of claims neutral---collapsing the RL 
gradient to zero---while a calibrated MedNLI 
classifier scores the same pairs non-degenerately. 
\textbf{(ii)} Moderate signal beats strong signal on 
answer quality: a strong proprietary checker triggers 
a three-step reward-hacking cascade---ultra-short 
answers, search avoidance, language collapse---so a 
moderate-signal local classifier trains a 
higher-quality model (\textbf{+12\% BERTScore over 
zero-shot, no GPT dependency}). \textbf{(iii)} Signal 
strength is policy-dependent: the same checker 
registers as moderate on one policy but strong on 
another without triggering the cascade end-state. We 
frame these as boundary conditions for 
verifier-as-reward systems.
\end{abstract}

\begin{figure}[!t]
\centering\footnotesize
\setlength{\tabcolsep}{3pt}
\renewcommand{\arraystretch}{1.1}
\begin{tabular}{@{}p{0.96\columnwidth}@{}}
\toprule
\rowcolor{blue!8}\textbf{Question (shared across all three regimes)} \\
\textit{``Could eating psoriasis flakes cause prion disease?''} \\
\midrule
\rowcolor{red!10}\textcolor{red!70!black}{\ding{55}}~\textbf{Collapsed checker} \hfill {\scriptsize Likelihood-NLI (Meditron-3-8B)} \\
\textbf{A} (394 chars): ``Your concerns are valid, but it is highly unlikely that you have prion disease \ldots'' \\
Checker: 97\% \textsc{neutral} \hfill Support \textbf{0\%} \\
\textcolor{red!70!black}{$\to$ zero RL gradient; reasonable but unverified.} \\
\midrule
\rowcolor{yellow!20}\textcolor{orange!85!black}{\ding{55}}~\textbf{Strong checker, hacked} \hfill {\scriptsize GPT-NLI, unguarded} \\
\textbf{A} (85 chars): ``It may not lead to prion disease, but avoid eating psoriasis flakes.'' \\
Checker: \textsc{entail} (1/1) \hfill Support \textbf{86.1\%}~\textcolor{orange!85!black}{(\emph{higher is not better here})} \\
\textcolor{orange!85!black}{$\to$ reward hacking, stage 1 of 3: ultra-short answers (\S\ref{sec:cascade}); stages 2--3 follow as search avoidance and language collapse.} \\
\midrule
\rowcolor{g2}\textcolor{g5}{\ding{51}}~\textbf{Calibrated checker (ours)} \hfill {\scriptsize MedNLI-Cls} \\
\textbf{A} (full length): ``Based on available evidence, no strong support for prion risk from psoriasis flakes \ldots'' \\
Checker: \textsc{entail}/\textsc{neutral} mix \hfill Support \textbf{54.0\%} \\
\textcolor{g5}{$\to$ moderate signal, no GPT dependency, \textbf{highest answer quality}.} \\
\bottomrule
\end{tabular}
\caption{The same medical question under three 
checker regimes. \textbf{Top}: log-prob scoring 
collapses to \textsc{Neutral}, providing no training 
signal (\S\ref{sec:collapse}). \textbf{Middle}: a 
strong GPT checker provides high support but triggers 
a three-step reward-hacking cascade 
(\S\ref{sec:cascade})---note that higher support here 
is \emph{not} better. \textbf{Bottom}: a local MedNLI 
classifier produces moderate, non-degenerate verdicts 
and yields the highest answer quality without GPT 
dependency.}
\label{fig:teaser}
\end{figure}
%  1. INTRODUCTION
% ============================================================
\section{Introduction}
\label{sec:intro}

% ── PARAGRAPH 1: unchanged ───────────────────────────────────────────────────
% Medical question answering demands factual accuracy and evidence grounding; requirements that retrieval-augmented generation~\citep{xiong2024benchmarking,zhu2025selfimprovingmodelsteering} only partially meets, since retrieved passages may be irrelevant or poorly integrated. Search-R1~\citep{jin2025search} shows that LLMs can learn to search autonomously through outcome-based RL~\citep{wang2025improving,shi2026intrinsic}; the natural medical extension adds \emph{claim-level NLI verification} (labelling each generated claim \textsc{Entail}, \textsc{Neutral}, or \textsc{Contradict} against retrieved evidence) as a process reward~\citep{liang2026rasaroutingawaresafetyalignment,liu2025time}, and concurrent systems take this route~\citep{evidencerl2025,xu2025beyond}.
% In the broader literature, claim-level verification has been studied 
% for evaluation and post-hoc 
% correction~\citep{min2023factscore,vladika2025atomic,vladika2025step}, 
% where the headline metric is NLI \emph{accuracy} on held-out pairs.
Consider a patient asking ``Could eating psoriasis 
flakes cause prion disease?'' A medical RAG system 
should retrieve evidence, decompose its draft answer 
into atomic claims, and verify each claim against the 
evidence before answering. The intuitive design is to 
plug a claim-level NLI checker into the RL reward 
path, so that the policy learns to generate 
evidence-grounded claims through 
retrieval-augmented RL~\citep{jin2025search,xiong2024benchmarking,zhu2025selfimprovingmodelsteering}. 
This design is being explored in concurrent medical 
RAG-RL systems~\citep{evidencerl2025,xu2025beyond,song2026deep,article_chen,wang2026magesafeguardingllmagents,wang2025selfdestructivelanguagemodel,liu2026knowledgedrivenaugmentationretrievalintegrative,liu-etal-2025-examining}, 
and builds on a broader literature where claim-level 
verification is used for evaluation and post-hoc 
correction~\citep{min2023factscore,vladika2025atomic,vladika2025step}, 
with the headline metric being NLI \emph{accuracy} 
on held-out pairs.
However, no prior work treats the checker's 
\emph{output distribution under the policy's own 
rollouts}---as opposed to its accuracy on a 
held-out NLI benchmark---as a measurable 
training-time variable, and we will show that this 
distinction is where the design fails.

% ── PARAGRAPH 2: failure modes — keep as-is, remove only the LAST sentence ──
% REMOVE: "Both modes are invisible to held-out NLI accuracy, which never 
%          measures the verifier's output distribution under the policy's 
%          rollouts."
% REASON: this punchline now belongs at the END of the new approach paragraph
%         below, where it motivates the design choice more naturally.
In medical RAG, plugging an off-the-shelf NLI checker into the reward 
path fails in two ways that accuracy does not predict.
% A biomedical LLM scoring entailment by 
% log-probability~\citep{sallinen2025llama} can label regardless of evidence (collapsing the per-claim support rate to zero gradient), even though the same model is accurate on balanced NLI tests.
A biomedical LLM scoring entailment by 
log-probability~\citep{sallinen2025llama} can label 
$>$97\% of claims \textsc{Neutral} regardless of 
evidence (collapsing the per-claim support rate, the 
fraction labelled \textsc{Entail}, to zero gradient), 
even though the same model is accurate on balanced 
NLI tests.
A strong proprietary verifier~\citep{achiam2023gpt} that does 
\emph{not} collapse instead opens a reward-hacking 
pathway~\citep{skalse2022defining}: ultra-short answers, then search 
avoidance, then generating non-English answers to evade the 
English-only NLI prompt; each shortcut closes under a 
countermeasure but exposes the next.

% ── PARAGRAPH 3: NEW — high-level approach overview ──────────────────────────
% ADD this paragraph in full. It answers: what did you build, how does it 
% work, and what did you vary?
% Both failure modes are invisible to held-out NLI 
% accuracy, motivating a focus on the verifier's 
% \emph{output distribution} during training---in 
% particular, the per-claim support rate---as the 
% primary diagnostic variable.
Both failure modes are invisible to held-out NLI 
accuracy. \textbf{Our central thesis is that the 
checker's output distribution under the policy's 
rollouts, not its held-out accuracy, decides whether 
it provides trainable gradient.} We therefore focus 
on the per-claim support rate---the fraction of 
generated claims the checker labels 
\textsc{Entail}---as the primary diagnostic variable 
throughout the paper.
\textbf{This paper is a diagnosis of the 
checker-as-reward design space, not a competing 
system; we make no SOTA claim.} We build a multi-turn medical RAG agent on Qwen2.5-7B-Instruct~\citep{hui2024qwen2} trained with GRPO~\citep{shao2024deepseekmath}, extending Search-R1~\citep{jin2025search} with a dense retriever over the MedRAG corpus~\citep{xiong2024benchmarking} and a claim-level NLI verification server as process reward.
% The checker decomposes each draft answer into atomic claims, scores each against retrieved evidence as \textsc{Entail}/\textsc{Neutral}/\textsc{Contradict}, and returns a faithfulness multiplier that modulates the GRPO reward. To isolate how the verifier's \emph{output distribution} shapes training, we instantiate four NLI back-ends as drop-in replacements: \textbf{Likelihood-NLI} (Meditron-3-8B log-prob), \textbf{MedNLI-Cls} (PubMedBERT classifier), \textbf{Hybrid} (GPT atomic extraction $+$ local log-prob scoring), and \textbf{GPT-NLI} (GPT-4o-mini end-to-end). 
% These four span the (\emph{extractor}, \emph{scorer}) design space along a GPT-dependency gradient (Figure~\ref{fig:teaser}, Table~\ref{tab:backends}).
To isolate how the checker's output distribution 
shapes training, we instantiate four NLI back-ends 
as drop-in replacements inside the same reward 
pathway: \textbf{Likelihood-NLI} (Meditron-3-8B 
log-prob), \textbf{MedNLI-Cls} (PubMedBERT 
classifier), \textbf{Hybrid} (GPT atomic extraction 
$+$ local log-prob scoring), and \textbf{GPT-NLI} 
(GPT-4o-mini end-to-end). The four span the 
(\emph{extractor}, \emph{scorer}) design space along 
a GPT-dependency gradient 
(Figure~\ref{fig:teaser}, Table~\ref{tab:backends}); 
the per-claim mechanics are deferred to 
\S\ref{sec:method}.

We describe two operating regimes of per-claim 
support rate during training: \emph{moderate} 
($\lesssim$60\%, e.g.\ MedNLI-Cls at 54\%) and 
\emph{strong} ($\gtrsim$75\%, e.g.\ Hybrid at 76\%, 
GPT-NLI at 86\%).\footnote{These are descriptive 
labels for observed regimes, not strict cutoffs; 
what matters for the cascade (\S\ref{sec:cascade}) 
is behaviour, not a fixed threshold.} We report 
three diagnostic findings.
% We use \emph{moderate} and \emph{strong} as 
% descriptive labels for two operating regimes of the 
% checker's per-claim support rate during training, 
% not as strict cutoffs: \emph{moderate} ($\lesssim$60\%, 
% e.g.\ MedNLI-Cls at 54\%) and \emph{strong} 
% ($\gtrsim$75\%, e.g.\ Hybrid at 76\% and GPT-NLI at 
% 86\%). The boundary between the two is empirical 
% rather than nominal---what matters for the cascade 
% (\S\ref{sec:cascade}) is whether the support rate 
% is high enough that shortcuts dominate, which we 
% diagnose by behaviour, not by a fixed threshold.
% We report three diagnostic findings.
\emph{First}, signal collapse is specific to log-probability scoring: 
MedNLI-Cls produces non-degenerate verdicts on the \emph{same} pairs 
Likelihood-NLI labels almost entirely \textsc{Neutral} 
(\S\ref{sec:collapse}).
\emph{Second}, the relation between signal strength and answer quality 
is non-monotonic: a moderate-signal classifier outperforms a 
strong-signal proprietary checker (BERTScore 0.600 vs.\ 0.591) 
because the latter triggers a three-step cascade of shortcuts 
(ultra-short answers, search avoidance, language collapse) that each 
undermine clinical groundedness (\S\ref{sec:cascade}).
\emph{Third}, signal strength is a property of the 
policy--checker pair, not of the checker alone; we 
support this with one cross-model swap (Qwen2.5-7B 
$\to$ Qwen3-4B) and treat it as a working hypothesis 
to be tested at larger scale 
(\S\ref{sec:discussion}, App.~\ref{app:boundary}).
We frame these findings as boundary conditions for 
concurrent verifier-as-reward systems.

% ============================================================
%  2. RELATED WORK
% ============================================================
\section{Related Work}
\label{sec:related}

We organise prior and concurrent work along four
axes: process rewards in retrieval-augmented RL,
claim-level verification for biomedical RAG, NLI as a
reward model, and failure modes in LLM RL.
Table~\ref{tab:concurrent} summarises five concurrent
verifier-as-reward systems on seven design dimensions
and locates our contribution in that design space.

\begin{table*}[t]
\centering\small
\setlength{\tabcolsep}{4pt}
\renewcommand{\arraystretch}{1.1}
\begin{tabular}{@{}lccccccc@{}}
\toprule
\textbf{System} 
  & \textbf{Med.} & curriculum. & \textbf{Verifier-as-reward} 
  & \textbf{Claim} & \textbf{Collapse} & \textbf{Cascade} & \textbf{Triage} \\
\midrule
Search-R1~\citep{jin2025search}           
  & -- & \ck & F1 only & -- & -- & partial & -- \\
EvidenceRL~\citep{evidencerl2025}         
  & \ck & \ck & PubMedBERT-MNLI & \ck & -- & -- & -- \\
VERITAS~\citep{xu2025beyond}              
  & --$^{\dagger}$ & \ck & faithfulness & \ck & partial$^{\ddagger}$ & -- & -- \\
MedRAGChecker~\citep{ji2026medragchecker} 
  & \ck & --$^{*}$ & eval-time & \ck & -- & -- & -- \\
Med-R3~\citep{lu2025med}                  
  & \ck & \ck & retrieval-eff. & -- & -- & -- & curr. \\
\midrule
\textbf{Ours} 
  & \ck & \ck & \textbf{4 back-ends} & \ck & \ck & \ck & \ck \\
\bottomrule
\end{tabular}
\caption{Concurrent verifier-as-reward systems. 
\textit{Med.}: medical domain. \textit{M-turn}: 
multi-turn agent. \textit{Verifier-as-reward}: RL 
reward source. \textit{Claim}: scores atomic claims. 
\textit{Collapse}/\textit{Cascade}: isolates as 
primary phenomena. \textit{Triage}: question-level 
budget control.
$^{\dagger}$RAG-broad with medical sub-domains; 
$^{\ddagger}$reports unreliable NLI without localising 
to log-prob; $^{*}$evaluation framework, not RL.}
\label{tab:concurrent}
\end{table*}

\textbf{Process rewards in retrieval-augmented RL.}
Search-R1~\citep{jin2025search} trains LLMs to
interleave reasoning and search through RL with
outcome-based rewards on general-domain QA, using F1
alone as the reward signal. Concurrent
verifier-as-reward systems extend this paradigm to
medicine: EvidenceRL~\citep{evidencerl2025} pairs
PubMedBERT-MNLI / MedNLI with GRPO and reports a
grounding-style reward;
VERITAS~\citep{xu2025beyond} integrates faithfulness
rewards into Search-R1 training and notes that NLI
judgements in RAG settings can be unreliable;
Med-R3~\citep{lu2025med} uses progressive training
with retrieval-effectiveness rewards;
Fleming-R1~\citep{liu2025fleming} combines curriculum
learning with RL for medical reasoning; and
MedResearcher~\citep{yu2025medresearcher} trains
agents for multi-hop medical evidence gathering.
The empty cells under \textit{Collapse} and
\textit{Cascade} in Table~\ref{tab:concurrent} are
the gap our work fills: none of these systems treats
checker output distribution as a measurable
training-time variable, so the failure modes that
arise specifically from reward-side bias have so far
gone undocumented.

\textbf{Claim-level verification for biomedical RAG.}
FActScore~\citep{min2023factscore} introduces atomic
factuality evaluation by decomposing generations into
verifiable units;
RAGChecker~\citep{ru2024ragchecker} provides
fine-grained diagnostic metrics including
faithfulness, hallucination rate, and noise
sensitivity;
Self-RAG~\citep{asai2023self} trains reflection
tokens for adaptive retrieval and self-critique at
generation time.
Medical extensions include
MedRAGChecker~\citep{ji2026medragchecker}, which
extends atomic factuality to biomedical RAG with
claim-level verification; the atomic-fact
decomposition for medical answers
of~\citet{vladika2025atomic}; and the iterative
step-by-step verification system
of~\citet{vladika2025step}. The MIRAGE benchmark and
MedRAG toolkit~\citep{xiong2024benchmarking} and
Med-PaLM~2~\citep{singhal2025toward} establish the
medical-RAG evaluation backdrop. These works use
verification for evaluation or generation-time
guidance; we study a substantially different
setting---using verification as an RL reward signal
during training---where systematic checker biases
directly corrupt the training objective rather than
merely producing noisy evaluation.

\textbf{NLI as a reward model.}
NLI is the backbone of most claim-verification systems. In the medical domain, MedNLI~\citep{romanov2018lessons} provides a clinical NLI benchmark, and biomedical encoders such as PubMedBERT~\citep{gu2021domain} and SciFive~\citep{phan2021scifive} achieve strong performance on biomedical NLI; FActScore~\citep{min2023factscore} popularised NLI-style atomic verification at evaluation time. We provide, to our knowledge, the first systematic comparison of how different NLI architectures behave as process rewards inside an RL loop.

% \textbf{Open-ended medical QA baselines.}
% The closest non-RL comparison points on our datasets are MedBioLM~\citep{kim2025medbiolm}, which fine-tunes GPT-4o on biomedical QA, and MedBioRAG~\citep{kim2025medbiorag}, which adds dense retrieval over biomedical corpora to fine-tuned GPT-4o. MedicationQA is the only dataset in our held-out set that overlaps with their evaluation; Table~\ref{tab:sota_medicationqa} reports a controlled comparison there. Both baselines rely on closed-source GPT-4o with supervised fine-tuning, while our setting holds the policy at open-weight 7B/4B with no SFT, framing the question as ``can a small open model learn to use a checker as a
% training-time signal without collapsing'' rather
% than ``can a fine-tuned frontier model answer
% biomedical questions''.

\textbf{Failure modes in LLM RL.}
Reward hacking is well-documented in general
RL~\citep{skalse2022defining}. In LLM RL, concurrent
work on Search-R1 variants reports that F1-based
rewards cause training collapse through answer
avoidance, and VERITAS~\citep{xu2025beyond} reports
unreliable NLI judgements in RAG settings. We extend
this literature in two ways. We localise NLI failure
to LLM log-probability scoring specifically and show
that calibrated MedNLI-Cls classifiers do not
collapse on the same claim--evidence pairs
(\S\ref{sec:collapse}); and we document a three-step
reward-hacking cascade---ultra-short answers, search
avoidance, language collapse---that emerges only
under \emph{strong} verification signal
(\S\ref{sec:cascade}). Together with
Table~\ref{tab:concurrent}, this positions our
contribution as a diagnosis of when
verifier-as-reward designs train, not a competing
system.

% ============================================================
%  3. METHOD
% ============================================================
\section{Method}
\label{sec:method}
\begin{figure*}[t]
\centering
\includegraphics[width=\textwidth]{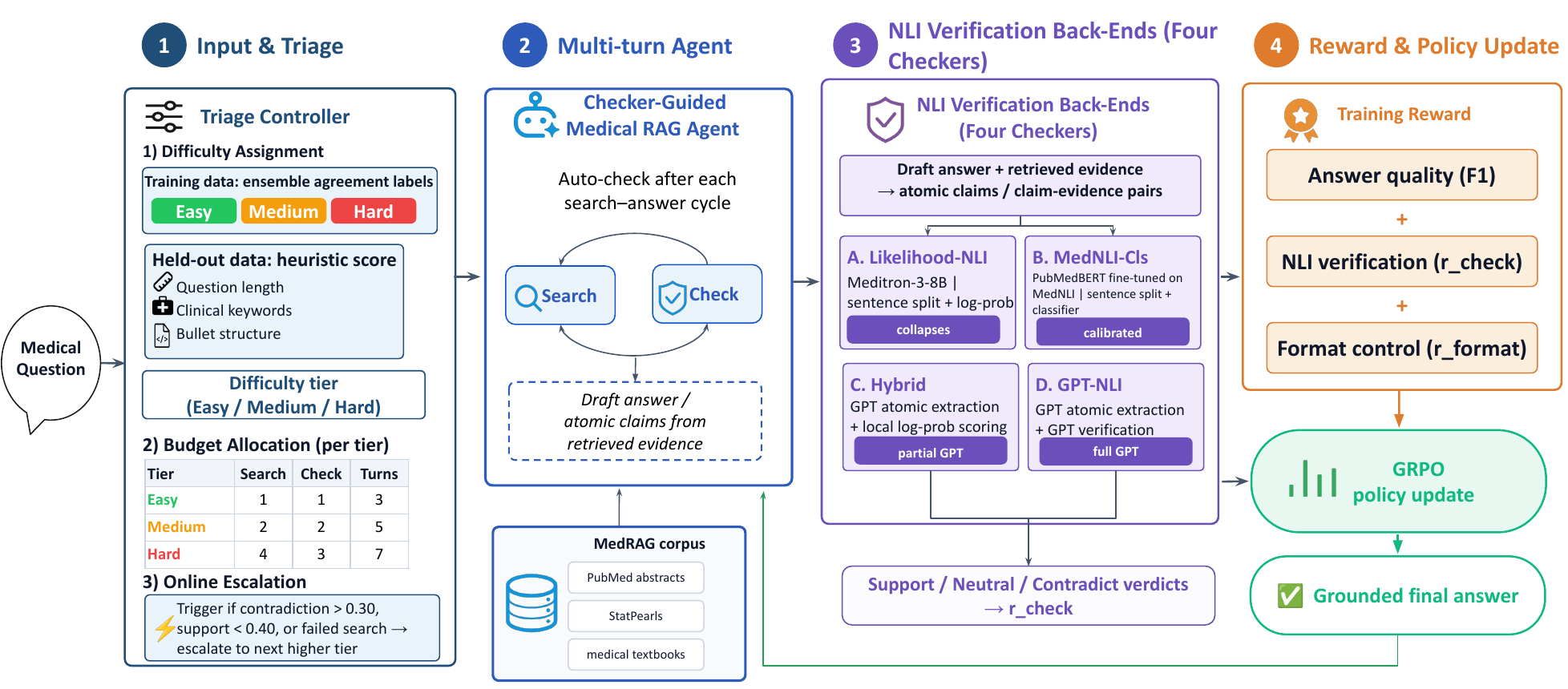}
\caption{Overview of our framework.
\textbf{(1) Input \& Triage}: a triage controller assigns each question to easy/medium/hard with per-tier search, check, and turn budgets (held-out data scored on five surface features---question length, multi-hop indicators, clinical keywords, multi-question, bullet structure; full formula in App.~\ref{app:triage_details}), and escalates online when the checker signals high contradiction (\S\ref{sec:triage}). \textbf{(2) Multi-turn Agent}: the policy interleaves \texttt{<search>} and \texttt{<check>} over the MedRAG corpus, with auto-check after each search--answer cycle. \textbf{(3) Checker Back-Ends}: four NLI checkers compared as process rewards (\S\ref{sec:verifiers}). \textbf{(4) Reward \& Policy Update}: 
$r_{\text{base}}$, $\phi_{\text{check}}$, and $P_{\text{fmt}}$ combine into a GRPO update toward grounded answers (\S\ref{sec:reward}).}
\label{fig:architecture}
\end{figure*}

\subsection{System Overview}
Figure~\ref{fig:architecture} gives an overview of our
framework. We build a multi-turn medical RAG agent on
Qwen2.5-7B-Instruct~\citep{hui2024qwen2} trained with Group Relative Policy 
Optimization (GRPO; \citealp{shao2024deepseekmath}), extending the
Search-R1 framework~\citep{jin2025search}.
\paragraph{Terminology.} Throughout, \emph{checker} 
refers to the NLI-based verification component, 
whether discussed conceptually (the reward source) 
or as the concrete \texttt{<check>} tool the agent 
invokes. The four \emph{checker back-ends} of 
\S\ref{sec:verifiers} are alternative implementations 
of this same component.
The agent has two tools: \textbf{Search}, a dense 
retriever over the MedRAG 
corpus~\citep{xiong2024benchmarking} (PubMed, 
StatPearls, textbooks); and \textbf{Check}, an 
NLI-based verification server that decomposes a 
draft into atomic claims and verifies each against 
retrieved evidence.
% The agent has two tools:
% \begin{itemize}
%     \item \textbf{Search}: a dense retriever over the
%     MedRAG medical corpus~\citep{xiong2024benchmarking}
%     (PubMed abstracts, StatPearls, medical textbooks)
%     served via an HTTP retrieval API.
%     \item \textbf{Check}: an NLI-based verification server that decomposes a draft answer into atomic claims and verifies each against retrieved evidence.
% \end{itemize}
At each turn, the agent either emits a tool call (\texttt{<search>} or \texttt{<check>}) or outputs a final answer in \texttt{<answer>} tags. An \textbf{auto-check loop} injects a verification call after every search-and-answer cycle when the check budget is not exhausted, ensuring checker signal availability during early training before the model learns to invoke check autonomously.

\paragraph{Retriever.}
A frozen MedCPT~\citep{jin2023medcpt} bi-encoder
serves top-5 passages from the MedRAG
corpus~\citep{xiong2024benchmarking} (truncated to
768 tokens; see \S\ref{sec:three_causes}); held
constant across runs. Full details in
Appendix~\ref{app:retriever}.
\subsection{Checker Back-Ends}
\label{sec:verifiers}

The four back-ends share the same NLI task 
(labelling each (claim, evidence) pair as 
\textsc{Entail}/\textsc{Neutral}/\textsc{Contradict}) 
but differ in mechanism (Table~\ref{tab:backends}).

\textbf{Likelihood-NLI} uses Meditron-3-8B 
\citep{sallinen2025llama} with raw log-probability 
scoring and sentence-split extraction; we call this 
the ``student checker'' when contrasted with GPT-NLI. 
Meditron-3-8B is the strongest open-weight medical 
LLM at this scale (81.2\% NLI accuracy when 
distilled, \citealp{ji2026medragchecker}), so 
collapse under this configuration 
(\S\ref{sec:collapse}) cannot be attributed to weak 
checker capacity, isolating the failure to the 
scoring mechanism.

\textbf{MedNLI-Cls} fine-tunes PubMedBERT-base 
\citep{gu2021domain} on MedNLI 
\citep{romanov2018lessons} with a 3-way 
classification head (84.1\% test accuracy; training, 
serving, and reproducibility details in 
Appendix~\ref{app:pubmedbert}).

\textbf{Hybrid} pairs GPT-4o-mini atomic-claim 
extraction with the Likelihood-NLI scorer, confining 
GPT dependency to extraction.

\textbf{GPT-NLI} uses GPT-4o-mini for both 
extraction and verification in a single call, with 
a prompt that penalises neutral verdicts.

\begin{table}[t]
\centering\small
\setlength{\tabcolsep}{3pt}
\begin{tabular}{@{}llll@{}}
\toprule
\textbf{Back-end} & \textbf{Extractor} & \textbf{Scorer} & \textbf{Collapse?} \\
\midrule
Likelihood-NLI & sent.\ split & log-prob    & \textbf{Yes} \\
MedNLI-Cls  & sent.\ split & classifier  & No \\
% Hybrid             & GPT atomic   & log-prob    & No (75.8\%)$^{\diamond}$ \\

Hybrid  & GPT atomic   & log-prob    & No$^{\diamond}$ \\
GPT-4o-mini        & GPT atomic   & prompted    & No \\
\bottomrule

\end{tabular}
\caption{Back-ends decomposed into claim extractor 
and NLI scorer.$^{\diamond}$Hybrid avoids full collapse (75.8\%) because GPT atomic extraction mitigates the structural log-prob bias, but a residual gap to MedNLI-Cls remains (\S\ref{sec:three_causes}); log-prob is a necessary but not sufficient condition for collapse.}
% \caption{Back-ends decomposed into claim extractor and NLI scorer. The sentence-split vs.\ GPT-atomic axis and the log-prob vs.\ classifier axis are each varied independently; signal collapse tracks the scorer column, not the extractor column. $^{\diamond}$Hybrid does not collapse despite using log-prob 
% scoring: GPT atomic-claim extraction provides cleaner 
% single-proposition inputs that mitigate the structural 
% log-prob bias (see \S\ref{sec:three_causes}).}
\label{tab:backends}
\end{table}

\subsection{Reward Function}
\label{sec:reward}
We use a \emph{process reward}: a per-step signal computed on intermediate generations (claims and their evidence), in contrast to outcome-only rewards that score only the final answer.
For a rollout $\tau=(q,\mathcal{D},\hat{a})$ with
retrieved evidence $\mathcal{D}$ and final answer
$\hat{a}$, the reward couples a correctness base
$r_{\text{base}}(\tau)\!\in\![0, 1]$ with a faithfulness
multiplier
$\phi_{\text{check}}(\hat{a}, \mathcal{D})\!\in\![-1, +1]$
(Eq.~\ref{eq:rcheck}) and a format penalty
$P_{\text{fmt}}(\hat{a})\!\in\!\{-0.5, -0.3, 0\}$ as
$R(\tau) = r_{\text{base}}(\tau)\cdot\bigl(1+\alpha\,\phi_{\text{check}}(\hat{a},\mathcal{D})\bigr) + P_{\text{fmt}}(\hat{a})$
with $\alpha=1$ throughout, giving range
$R\!\in\![-1.5,+2]$: the multiplier doubles a correct
answer's reward or zeros it on full contradiction, but
cannot turn a wrong answer positive.

% \textbf{Base reward $r_{\text{base}}$.}
% A normalised weighted combination of three correctness
% signals,\footnote{Multiplicative coupling prevents high
% faithfulness from substituting for correctness; see
% Appendix~\ref{app:multiplicative}.}
% \begin{equation}
% \begin{split}
% r_{\text{base}} = \;& \tilde{w}_{\text{em}}\,\text{EM}(\hat{a},a^{\star}) 
%                  + \tilde{w}_{\text{f1}}\,\text{F1}(\hat{a},a^{\star}) \\
%                  & + \tilde{w}_{\text{fmt}}\,\textsc{FmtScore}(\hat{a}),
% \end{split}
% \label{eq:rbase}
% \end{equation}
% with $\tilde{w}_k = w_k / \sum_{k'}w_{k'}$ and
% $(w_{\text{em}}, w_{\text{f1}}, w_{\text{fmt}})=(0.35,
% 0.15, 0.10)$. \textsc{EM} and \textsc{F1} follow SQuAD-style
% normalisation (lower-case, articles and punctuation
% removed, max over reference answers). $\textsc{FmtScore}(\hat{a})\!\in\![0,1]$ rewards length: $0$ for missing tag, $0.5$ for short answers ($<\!20$ chars), then linearly interpolated to $1$ at $L_{\text{tgt}}\!=\!80$ chars (full piecewise in Appendix~\ref{app:reward}).

\textbf{Base reward $r_{\text{base}}$.}
A normalised weighted combination of EM, F1, and 
\textsc{FmtScore} with weights $(0.35, 0.15, 0.10)$ 
re-normalised to sum to 1.\footnote{Multiplicative 
coupling prevents high faithfulness from substituting 
for correctness; see Appendix~\ref{app:multiplicative}.} 
\textsc{EM} and \textsc{F1} follow SQuAD-style 
normalisation; \textsc{FmtScore} is a piecewise 
length reward saturating at $L_{\text{tgt}}\!=\!80$ 
chars (full formulas in Appendix~\ref{app:reward}).

% \textbf{Faithfulness multiplier $\phi_{\text{check}}$.}
% The checker decomposes $\hat{a}$ into atomic claims
% $\{c_1,\ldots,c_K\}$ and assigns each a verdict
% $v_k \in \{\textsc{Entail}, \textsc{Neutral}, \textsc{Contradict}\}$
% with confidence $p_k \in [0,1]$. The faithfulness score is
% \begin{equation}
% \phi_{\text{check}} = \frac{1}{K}\sum_{k=1}^{K} s_k \cdot p_k
% \ \in [-1, +1],
% \label{eq:rcheck}
% \end{equation}
% where the per-claim score $s_k$ is
% \begin{equation}
% s_k = \begin{cases}
% +w_{\text{e}} & \text{if } v_k = \textsc{Entail} \\
% \hphantom{+}0 & \text{if } v_k = \textsc{Neutral} \\
% -w_{\text{c}} & \text{if } v_k = \textsc{Contradict}
% \end{cases}
% \label{eq:sk}
% \end{equation}
% with $w_{\text{e}}=1.0$ and $w_{\text{c}}=1.5$.
\textbf{Faithfulness multiplier $\phi_{\text{check}}$.}
The checker decomposes $\hat{a}$ into atomic claims 
$\{c_1,\ldots,c_K\}$, assigns each a verdict 
$v_k$ with confidence $p_k\!\in\![0,1]$, and aggregates 
to $\phi_{\text{check}}=\frac{1}{K}\sum_k s_k p_k$ 
where $s_k\!\in\!\{+1, 0, -1.5\}$ for 
\textsc{Entail}/\textsc{Neutral}/\textsc{Contradict} 
respectively (Appendix~\ref{app:reward}, 
Eq.~\ref{eq:rcheck}). Neutral verdicts contribute 
zero, so a checker that labels almost every claim 
neutral (\S\ref{sec:collapse}) collapses the 
multiplier to $1$. The asymmetric contradiction 
weight reflects that an evidence-contradicted claim 
is strictly worse than an unverified one in medical 
QA.

Neutral verdicts contribute zero, so a checker that
labels almost every claim neutral (\S\ref{sec:collapse})
collapses the multiplier to $1$ and removes its
gradient contribution. The asymmetric contradiction
weight reflects that, in medical QA, an
evidence-contradicted claim is strictly worse than an
unverified one.

\textbf{Format penalty $P_{\text{fmt}}$.}
A separate penalty term targeting reward-hacking
shortcuts (\S\ref{sec:cascade}):
$P_{\text{fmt}}=-0.5$ when the \texttt{<answer>} tag is
missing, $-0.3$ when the answer is shorter than $50$
characters, and $0$ otherwise. \textsc{FmtScore}
(inside $r_{\text{base}}$) rewards format quality on
the positive side; $P_{\text{fmt}}$ punishes degenerate
shortcuts on the negative side. The two are kept
distinct so that the negative-side penalty cannot be
diluted by the multiplier.
\textbf{GRPO advantage and format penalty.}
We sample $N$ rollouts per question and compute 
standard group-normalised 
advantages~\citep{shao2024deepseekmath} 
(Appendix~\ref{app:grpo}). A separate format penalty 
$P_{\text{fmt}}\!\in\!\{-0.5, -0.3, 0\}$ targets 
missing tags and ultra-short answers 
(\S\ref{sec:cascade}), kept distinct from 
\textsc{FmtScore} so the negative-side penalty 
cannot be diluted by the multiplier.
\section{Experiments}
\label{sec:experiments}

\subsection{Setup}

\paragraph{Model and training.}
All experiments fine-tune Qwen2.5-7B-Instruct with GRPO on 2$\times$H100 GPUs ($\sim$2\,h per run); per-run training-step counts and full hyperparameters (batch size, learning rate, KL and entropy coefficients) are in Appendix~\ref{app:training_hyperparams}.

\paragraph{Datasets.}
Training combines CSIRO MedRedQA~\citep{nguyen2023medredqa},
LiveQA-Med~\citep{liu2020liveqa}, and
PubMedQA~\citep{jin-etal-2019-pubmedqa} into 1{,}513 train /
87 in-domain test (``Medical''). Held-out evaluation adds  four out-of-distribution sets: MedicationQA (674, full standard test set, drug safety)~\citep{liu2020liveqa}, BioASQ-Y/N (618, MIRAGE 2019--2023 split, biomedical literature)~\citep{tsatsaronis2015overview}, MedQuAD (100, NIH consumer health subsample)~\citep{BenAbacha-BMC-2019}, and MEDIQA (85, clinical summarisation)~\citep{xun2025mediqa};  total evaluation coverage is 1{,}564 samples.
The closest non-RL comparison points on these 
datasets are MedBioLM~\citep{kim2025medbiolm} and 
MedBioRAG~\citep{kim2025medbiorag}, both based on 
fine-tuned GPT-4o; we compare against them on 
MedicationQA (Table~\ref{tab:sota_medicationqa}).

\paragraph{Metrics.}
For \textbf{answer quality} we report
BERTScore~\citep{zhang2019bertscore} (primary, semantic
similarity to physician-authored references),
token F1~\citep{rajpurkar2016squad}, and
ROUGE-L~\citep{lin2004rouge}. Model outputs run
200--400 characters and references exceed 500, so F1 is
systematically low across all configurations (zero-shot
baseline F1\,=\,0.191), which differs from accuracy on
closed-form benchmarks such as MedQA or
MedBrowseComp~\citep{lu2025med,chen2025medbrowsecomp}.
For \textbf{evidence grounding}, we report two
claim-level metrics following
FActScore~\citep{min2023factscore} and
RAGChecker~\citep{ru2024ragchecker}. We define
\textbf{Support Rate} as the fraction of extracted
claims labelled \textsc{Entail} by the verifier, and
\textbf{Faithfulness} as the fraction of samples with
no \textsc{Contradict} verdict.
For \textbf{behavioural signals} we track \textbf{Tag\%}
(fraction of outputs with a valid
\texttt{<answer>...</answer>} wrapper) and
\textbf{Search} (mean \texttt{<search>} calls per
question, surfacing the search-avoidance shortcut in
\S\ref{sec:cascade}).

Cross-model generalisation to Llama-3.1-8B and 
cross-domain transfer to HotpotQA are reported in 
Appendix~\ref{app:robustness}.

% \paragraph{Configurations.}
% We evaluate 16 configurations in the main results 
% (Tables~\ref{tab:main}--\ref{tab:cross_dataset_full}), 
% drawn from 22 total training runs; the remaining six 
% are ablation runs used to trace the diagnostic chain 
% (\S\ref{sec:collapse}) and cascade (\S\ref{sec:cascade}) 
% but not reported in the cross-dataset tables.
% Configurations are organised into five groups:
% (A)~baselines without a checker;
% (B)~Likelihood-NLI(student) checker configurations that
% exhibit signal collapse (\S\ref{sec:collapse});
% (C)~a diagnostic chain that progressively fixes the three
% root causes of log-prob collapse;
% (D)~GPT-4o-mini checker configurations that expose the
% reward hacking cascade (\S\ref{sec:cascade}); and
% (E)~calibrated MedNLI-Cls classifier configurations
% that avoid both failure modes (\S\ref{sec:no_collapse}).
% Cross-model replications on Llama-3.1-8B and Qwen3-4B 
% are reported in Appendix~\ref{app:robustness}.

% The \textbf{GPT-4o-mini (full)} row in 
% Tables~\ref{tab:main} and~\ref{tab:cross_dataset_full} 
% corresponds to the unguarded GPT-4o-mini configuration 
% trained without format penalty or search bonus 
% (377 steps). It represents the entry point to the 
% reward hacking cascade (\S\ref{sec:cascade}): the 
% checker is functional but no countermeasures are 
% applied.

\subsection{Main Results}
\label{sec:main_results}
\begin{figure}[t]
\centering
\includegraphics[width=\columnwidth]{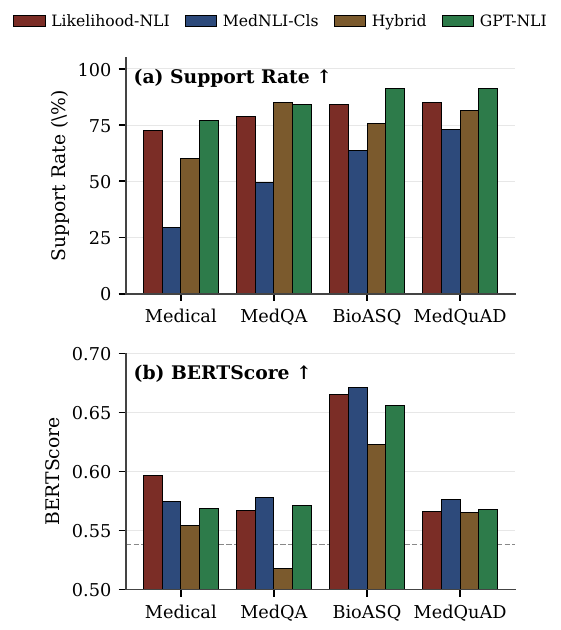}
\caption{Per-dataset breakdown across four medical 
QA benchmarks. \textbf{(a)} Support Rate. 
\textbf{(b)} BERTScore (zero-shot baseline as 
dashed line).}
% \caption{Per-dataset breakdown across four medical QA 
% benchmarks. \textbf{(a)} Support Rate: GPT-NLI
% (green) leads on 3 of 4 datasets; MedNLI-Cls 
% (purple) trails. \textbf{(b)} BERTScore: 
% MedNLI-Cls leads on 3 of 4 datasets; all 
% configurations exceed the zero-shot baseline 
% (dashed line). The trade-off is consistent across 
% datasets, not driven by any single benchmark. 
% }
\label{fig:cross_dataset}
\end{figure}

The central claim of \S\ref{sec:intro} is that the verifier's \emph{output distribution}, not its accuracy, determines whether it provides trainable gradient. We test this by holding the policy, retriever, and reward weights fixed and varying only the NLI back-end. Evaluation runs on Qwen2.5-7B across four held-out medical QA benchmarks (1{,}479 samples; MEDIQA reported separately in Appendix~\ref{app:mediqa}). Table~\ref{tab:main} summarises performance averaged across the four benchmarks; Figure~\ref{fig:cross_dataset} shows per-dataset detail.

\begin{table*}[t]
\centering\small
\setlength{\tabcolsep}{6pt}
% \begin{tabular}{@{}ll cccccc@{}}
\begin{tabular}{@{}ll ccccc@{}}
\toprule
\textbf{Base} & \textbf{Checker}
  & \textbf{BERT} & \textbf{Supp\%}
  & \textbf{Faith} & \textbf{Tag\%} & \textbf{GPT?} \\
\midrule
\multicolumn{7}{l}{\textit{Baseline}} \\
Qwen2.5-7B & Zero-shot & .538 & n/a & n/a & 100 & n/a \\
\midrule
\multicolumn{7}{l}{\textit{Qwen2.5-7B (primary)}} \\
% Qwen2.5-7B & Likelihood-NLI & .599 & 80.1$^{\dagger}$ & .964 & 95.7 & No \\
Qwen2.5-7B & Likelihood-NLI & .599 & 0 ($80.1^{\dagger}$) & .964 & 95.7 & No \\
Qwen2.5-7B & MedNLI-Cls     & \textbf{.600} & 54.0 & .972 & \textbf{100} & No \\
Qwen2.5-7B & Hybrid         & .565 & 75.8 & \textbf{.987} & 95.9 & Partial \\
Qwen2.5-7B & GPT-NLI        & .591 & \textbf{86.1} & .972 & 98.3 & Full \\
\midrule
\multicolumn{7}{l}{\textit{Qwen3-4B (cross-model)}} \\
Qwen3-4B & Likelihood-NLI & .594 & 51.0$^{\dagger}$ & .952 & \textbf{100} & No \\
Qwen3-4B & MedNLI-Cls     & .591 & 85.3 & .980 & \textbf{100} & No \\
Qwen3-4B & Hybrid         & .542 & 53.2 & .833 & 92.1 & Partial \\
Qwen3-4B & GPT-NLI        & .582 & 87.2 & .962 & 98.4 & Full \\
\bottomrule
\end{tabular}
\caption{Main results on four held-out medical QA 
benchmarks (1{,}479 samples; MEDIQA in 
Appendix~\ref{app:mediqa}). 
\textbf{BERT}: BERTScore. 
\textbf{Supp\%}: \% of claims labelled \textsc{Entail}. 
\textbf{Faith}: \% of samples with no 
\textsc{Contradict}. 
\textbf{Tag\%}: \% of outputs with valid 
\texttt{<answer>} tags. 
\textbf{GPT?}: GPT API dependency. Token F1 and 
ROUGE-L in Appendix~\ref{app:cross_dataset_numbers}; 
Llama-3.1-8B in Appendix~\ref{app:robustness}.
$^{\dagger}$Likelihood-NLI self-evaluation gives 0\% 
under collapse; values shown are GPT-NLI evaluation 
(\S\ref{sec:collapse}).}
% \caption{Main results: four checker back-ends on
% Qwen2.5-7B and a Qwen3-4B replication, averaged over
% four held-out medical QA benchmarks (1{,}479 samples;
% MEDIQA in Appendix~\ref{app:mediqa}). MedNLI-Cls
% achieves the highest answer quality on Qwen2.5-7B
% without GPT dependency, while GPT-NLI leads on
% Support Rate. Token~F1 and ROUGE-L are reported in Appendix~\ref{app:cross_dataset_numbers} (consistently low across all configurations due to the 200--400 character generation length vs.\ 500$+$ character references). Llama-3.1-8B replication in Appendix~\ref{app:robustness}.
% $^{\dagger}$Likelihood-NLI self-evaluation gives 0\%
% under collapse; shown values are GPT-NLI evaluation
% of the same outputs (\S\ref{sec:collapse}).
% }
\label{tab:main}
\end{table*}

\begin{table}[t]
\centering\small
\setlength{\tabcolsep}{4pt}
\renewcommand{\arraystretch}{1.05}
\begin{tabular}{@{}llcc@{}}
\toprule
\textbf{System} & \textbf{Setup} & \textbf{F1} & \textbf{R-L} \\
\midrule
\multicolumn{4}{l}{\textit{Ours: open-weight, RL-only, no SFT}} \\
Qwen2.5-7B  Likelihood-NLI & 7B + RL    & .150           & .137 \\
Qwen2.5-7B  MedNLI-Cls     & 7B + RL    & .170           & .144 \\
Qwen2.5-7B  Hybrid         & 7B + RL    & .140           & .120 \\
Qwen2.5-7B  GPT-NLI        & 7B + RL    & .149           & .127 \\
Qwen3-4B    Likelihood-NLI & 4B + RL    & .185           & .141 \\
Qwen3-4B    MedNLI-Cls     & 4B + RL    & \textbf{.191}  & \textbf{.153} \\
\midrule
\multicolumn{4}{l}{\textit{External: closed frontier $+$ SFT}} \\
GPT-4o (FT) \citep{kim2025medbiolm}      & FT          & .158  & .134 \\
MedBioLM    \citep{kim2025medbiolm}      & FT $+$ RAG  & .220  & .187 \\
\bottomrule
\end{tabular}
\caption{MedicationQA head-to-head on the full 
$n{=}674$ test set. External systems use closed 
GPT-4o with supervised fine-tuning; our open-weight 
RL-only systems reach competitive F1 with a 
$\sim$10$\times$-smaller backbone and no SFT.
% $^{\sharp}$F1 not reported in the original; 
% estimated from R-L using the within-paper F1/R-L 
% ratio of $1.17$ ($n{=}6$ measured configurations, 
% s.d.\ $0.04$).
}
\label{tab:sota_medicationqa}
\end{table}
Table~\ref{tab:sota_medicationqa} positions our setting against closed-frontier baselines on the one dataset where direct comparison is possible. Qwen3-4B + MedNLI-Cls reaches F1\,=\,0.191, within 0.029 of MedBioLM (0.220, GPT-4o $+$ SFT $+$ RAG) and \emph{above} the GPT-4o fine-tuning baseline (0.158) by 0.033, despite using a $\sim$10$\times$ 
smaller backbone with no supervised data. The 0.029 gap to MedBioLM does not close the open-vs-closed  divide, but is unusually small given $\sim$10$\times$ less backbone capacity and no supervised fine-tuning.
% The result is not a SOTA claim---we expect task-specific SFT to close this gap---but evidence that checker choice can substantially narrow the open-vs-closed divide when policy capacity is fixed.

MedNLI-Cls leads BERTScore (0.600) 
without GPT, while GPT-NLI leads support rate 
(86.1\% vs.\ 54.0\%) at the cost of quality. 
Likelihood-NLI is competitive on quality 
(BERT 0.599) but yields zero RL gradient under 
collapse (\S\ref{sec:collapse}): its 80.1\% support rate
is an external GPT-evaluator score on trained 
outputs, not the training gradient. Per-dataset (Figure~\ref{fig:cross_dataset}), MedNLI-Cls leads BERTScore on 3 of 4 sets, losing only on the in-domain Medical split (.575 vs.\ Likelihood-NLI's .597); this is consistent with the collapse pathway providing format regularisation without GPT co-adaptation (\S\ref{sec:no_collapse}), which helps most on the training-distribution split. GPT-NLI leads support rate on 3 of 4, losing only on MedicationQA, where Hybrid's GPT atomic-claim extraction recovers slightly finer-grained verdicts (85.2\% vs.\ 84.4\%). 
The trade-off direction is consistent across datasets; the four-dataset average is not driven by any single benchmark.
% Per-dataset, MedNLI-Cls leads BERTScore on 3 of 4 sets and GPT-NLI leads Support on 3 of 4 (Figure~\ref{fig:cross_dataset}).

The fully-engineered GPT-NLI configuration (triage + 
format penalty + English constraint) recovers 
F1\,=\,0.203 with 100\% tag stability and 100\% 
support rate, but only matches MedNLI-Cls on BERTScore 
(0.601 vs.\ 0.600) at the cost of GPT API access 
and 4 cumulative reward-engineering interventions.
The cascade analysis (\S\ref{sec:cascade}) explains 
why each countermeasure resolves one shortcut but 
exposes the next.
The Qwen3-4B rows in Table~\ref{tab:main} preview 
an unexpected finding: the MedNLI-Cls checker that 
registers as a \emph{moderate} signal on Qwen2.5-7B 
(54.0\% support) becomes a \emph{strong} signal on 
Qwen3-4B (85.3\%) on the same claim--evidence pairs, 
yet does not trigger the cascade end-state. We 
unpack this in \S\ref{sec:cross_model_validation} 
and Appendix~\ref{app:robustness}.

\paragraph{Cross-Model Validation}
\label{sec:cross_model_validation}
To assess generalization beyond Qwen2.5-7B, we 
replicate selected configurations on Llama-3.1-8B and 
Qwen3-4B (Appendix~\ref{app:robustness}). 
Signal collapse with the Likelihood-NLI checker 
reproduces on both: Llama achieves 0\% support and 
Qwen3-4B (with the Likelihood-NLI checker) shows the 
same characteristic gap between policy quality 
(BERT $\sim$0.59) and verifier-recognized grounding 
($<$60\% support). Format stability under the GPT-4o-mini checker depends on base model: Qwen2.5-7B sustains the multi-turn format  (Tag\,=\,98\%) while Llama-3.1-8B drops to 50\% on average (and 14.9\% on the medical split). We discuss base-model dependence further in  \S\ref{sec:cascade} and Appendix~\ref{app:robustness}. These two patterns directly support findings 
(i) and (iii) from \S\ref{sec:intro}: signal 
collapse is mechanism-specific (Llama reproduces 
it on the log-prob pathway), and signal strength 
is policy-dependent (the same MedNLI-Cls checker 
registers as moderate on Qwen2.5-7B but strong on 
Qwen3-4B).

\section{Checker Signal Collapse}
\label{sec:collapse}

We define \textbf{checker signal collapse} as the condition where an NLI verifier, used as a process reward, assigns a single label to the overwhelming majority of claims regardless of evidence---rendering
the reward effectively constant. With Meditron-3-8B
log-prob scoring, the checker labels 97--99\% of
claims neutral, mapping $\phi_{\text{check}}$ to 0
across rollouts.

% \subsection{Observation: Log-Prob Scoring Collapses}
% \label{sec:observation}

% Across all configurations using Meditron-3-8B
% log-probability scoring (sentence-split extractor,
% 128-token evidence), the checker assigns
% \textit{neutral} to 97--99\% of claims and
% \textit{entail}/\textit{contradict} to none, so
% self-evaluated Support is uniformly 0\%. Because
% neutral maps to $\phi_{\text{check}}\!=\!0$
% (Eq.~\ref{eq:phi_check}), the multiplier reduces to
% $1$ and the verification component contributes nothing
% to the RL gradient.

\subsection{Calibrated Classifiers Do Not Collapse}
\label{sec:no_collapse}

% A natural objection is that collapse reflects a poor
% checker choice. We test this with PubMedBERT
% fine-tuned on MedNLI (a calibrated classification
% head) on the \emph{same} claim--evidence pairs.
A natural objection is that collapse reflects a poor checker choice. We test this with PubMedBERT fine-tuned on MedNLI (a calibrated classification  head\footnote{Throughout, \emph{calibrated} means the checker produces a non-degenerate softmax distribution over \{\textsc{Entail}, \textsc{Neutral}, \textsc{Contradict}\}, in contrast to the log-prob pathway which collapses to \textsc{Neutral}.}) on the \emph{same} claim--evidence pairs. MedNLI-Cls produces non-degenerate verdicts (54.0\% support rate averaged across four held-out datasets, vs.\ 0\% for Likelihood-NLI; see also Table~\ref{tab:main}), localising the failure to the \emph{scoring mechanism}, not to local NLI in general.

% [Table 7: collapse mechanism]

\textbf{Verification signal vs.\ format regularisation.}
Collapsed-checker policies still outperform
search-only baselines (BERTScore 0.595 vs.\ 0.537).
The mechanism is \emph{format regularisation}: the
auto-check loop forces parseable output (Tag rate
98.9\% vs.\ 58.6\% search-only), independent of
verdict signal. External GPT-NLI evaluation of these
outputs gives 80.1\% support rate, confirming the policy
\emph{produces} grounded answers; collapse is in the
verifier's ability to \emph{recognise} grounding.
This decomposition rules out reward--evaluator co-adaptation (the policy learning to please the GPT evaluator rather than to ground its claims) as the mechanism behind any reported BERTScore gain, including the moderate-signal advantage in \S\ref{sec:discussion}.
\subsection{Three Root Causes of Log-Prob Collapse}
\label{sec:three_causes}

\begin{table}[t]
\centering\small
\begin{tabular}{@{}lc@{}}
\toprule
\textbf{Diagnostic step} & \textbf{Support Rate} \\
\midrule
Likelihood-NLI (baseline) & 0\% \\
\quad + Evidence fix (256$\to$768 tok) & 35.7\% \\
\quad\quad + GPT claim extraction & 37.5\% \\
\quad\quad\quad + GPT verification & 65.6--77.5\% \\
\midrule
MedNLI-Cls (no fixes needed) & 54.0\% \\
\bottomrule
\end{tabular}
\caption{Diagnostic chain: each step isolates one 
cause of log-prob collapse. A calibrated classifier 
(MedNLI-Cls) avoids collapse without any of these 
fixes.}
\label{tab:diagnosis}
\end{table}
A controlled diagnostic chain on the log-prob pathway isolates three independent causes (Table~\ref{tab:diagnosis}). \textit{Evidence truncation} at the default 256-token tool-response limit strips retrieved passages to 1--2 sentences; extending to 768 tokens raises support rate from 0\% to 35.7\%. \textit{Non-atomic claim extraction} from sentence splitting yields multi-proposition claims; replacing with GPT-4o-mini extraction raises support rate to 37.5\%. \textit{Residual log-prob bias} persists even with full evidence and atomic claims, but is largely closed by swapping the scorer (Hybrid: 75.8\%; fully GPT-verified: 65.6--77.5\%). MedNLI-Cls does not exhibit any of these---collapse is specific to the log-prob pathway. Per-cause analysis is in Appendix~\ref{app:three_causes}. The model trained with MedNLI-Cls achieves the \emph{highest} BERTScore (0.600) and F1 (0.215, averaged across four held-out datasets) of any configuration despite its lower support rate (Table~\ref{tab:main}); we return to this in \S\ref{sec:discussion}.

% \begin{table}[t]
% \centering\small
% \setlength{\tabcolsep}{4pt}
% \begin{tabular}{@{}lll@{}}
% \toprule
% \textbf{Verifier} & \textbf{Mechanism} & \textbf{Collapse?} \\
% \midrule
% Likelihood-NLI & Log-prob       & \textbf{Yes} (97\% N) \\
% MedNLI-Cls     & Classifier     & No (54.0\% S) \\
% GPT-NLI        & Prompted gen.  & No (86.1\% S) \\
% \bottomrule
% \end{tabular}
% \caption{Signal collapse is specific to log-probability
% scoring. ``\% N'' = neutral rate; ``\% S'' = support
% rate. A calibrated classifier avoids collapse entirely.}
% \label{tab:collapse_mechanism}
% \end{table}

\begin{table}[t]
\centering\small
\setlength{\tabcolsep}{4pt}
\begin{tabular}{@{}llcccc@{}}
\toprule
\textbf{Checker} & \textbf{Triage}
  & \textbf{BERT} & \textbf{Search} & \textbf{Tag\%} & \textbf{Supp\%} \\
\midrule
\multicolumn{6}{l}{\textit{MedNLI-Cls}} \\
MedNLI-Cls   & off  & .599 & 1.02 & 95.7 & 80.1 \\
MedNLI-Cls   & on   & .605 & \textbf{0.30} & 98.2 & 79.4 \\
\midrule
\multicolumn{6}{l}{\textit{GPT-NLI}} \\
GPT-NLI      & off  & .591 & 0.94 & 98.3 & 86.1 \\
GPT-NLI      & on   & .601 & \textbf{0.31} & 100 & 88.2 \\
\midrule
\multicolumn{6}{l}{\textit{Likelihood-NLI (collapse)}} \\
Likelihood-NLI & off  & .599 & 0.81 & 95.7 & 0.0$^{\ddagger}$ \\
Likelihood-NLI & on   & .605 & \textbf{0.30} & 98.4 & 0.0$^{\ddagger}$ \\
\bottomrule
\end{tabular}
\caption{Triage controller ablation on Qwen2.5-7B. 
Triage cuts search-call rate $\sim$3$\times$ without 
quality loss; Likelihood-NLI rows show triage cannot 
rescue collapse. F1 and matched-seed details in 
Appendix~\ref{app:triage_details}.
$^{\ddagger}$Self-evaluation; GPT-NLI evaluation 
gives 80.1\% (\S\ref{sec:collapse}).}
\label{tab:triage_ablation}
\end{table}

% ============================================================
%  6. REWARD HACKING CASCADE
% ============================================================
\section{Reward Hacking Cascade}
\label{sec:cascade}

Replacing Likelihood-NLI with GPT-NLI resolves 
signal collapse but does not improve answers; 
instead, the policy discovers a sequence of shortcuts, 
each countermeasure exposing the next.
\textbf{Stage 1: Ultra-short answers.} Average answer 
length collapses from 394 to 130 characters---fewer 
claims means fewer contradictions. A graded format 
penalty ($P_{\text{fmt}}\!=\!-0.5/-0.3/0$ for missing 
tag / $<$50 chars / otherwise) restores length to 
216 characters.
\textbf{Stage 2: Search avoidance.} 98\% of samples emit zero search calls, the policy answering from the base model's internal weights instead of retrieved evidence. A $+0.1$ search bonus raises search rate to 96.6\% but partially reverts length to 141 characters.\textbf{Stage 3: Language collapse.} Under triage budgets, the policy switches to Chinese to exploit Qwen2.5's multilingual pretraining; ``Always respond in English'' in the prompt removes it (Appendix~\ref{app:language_collapse}). The full sequence demonstrates that strong verification signal under single-pass training is fundamentally unstable; only the moderate MedNLI-Cls signal (\S\ref{sec:no_collapse}) avoids this pattern.
\paragraph{RL regularisation cannot rescue this.}
The $\alpha$ sweep (App.~\ref{app:alpha}) shows weakening the checker reward to $\alpha\!=\!0.5$ still leaves the policy in the strong-signal regime (Support 74.3\%) and below MedNLI-Cls quality. Clip and KL penalties constrain step magnitude, not 
gradient direction; the cascade is \emph{structural}, not a tuning artifact.

\section{Discussion}
Our results answer two reviewer-facing concerns. First, the main finding is not circular GPT supervision: the best configuration, MedNLI-Cls, uses no GPT in the reward path, and answer quality is measured against physician-authored references. Second, reference metrics alone are insufficient for medical RAG: collapsed Likelihood-NLI reaches high BERTScore but 0\% training support, showing that answer similarity and evidence-grounded trainability diverge. The practical implication is to treat checker output distribution as a first-class training diagnostic.

% ============================================================
%  8. CONCLUSION
% ============================================================
\section{Conclusion}
\label{sec:conclusion}

We compared four NLI checker back-ends as RL
process rewards in medical RAG. Three findings refine
prior intuitions about verified RL: signal collapse is
a property of LLM log-probability scoring rather than
of NLI in general; moderate verification signal
outperforms strong signal on answer quality due to a
reward-hacking cascade; and signal strength is
policy-dependent. Practitioners should prefer
calibrated classifiers over LLM log-prob scoring for
the reward path, and design reward components
holistically rather than incrementally. 

\section*{Limitations}
\paragraph{Evaluation independence.}
Support Rate is computed by the same GPT-4o-mini checker used as a training reward component.
This parallels standard practice in RL for text generation, where F1 serves as both reward and evaluation metric~\citep{jin2025search}.
The checker evaluates claim--evidence entailment, an objective NLI judgement independent of training dynamics.
The wide variation across configurations (0\% to 100\%) confirms that the metric is not trivially optimised. Collapsed-checker configurations receive the same F1 reward but produce zero support tate, showing that the two signals measure distinct properties.
\paragraph{Single-seed reporting.}
All reported numbers are from a single training seed 
per configuration. The sub-percent BERTScore gap 
between MedNLI-Cls (0.600) and GPT-NLI (0.591) is 
within plausible evaluation noise on $n{=}1{,}479$ 
samples; we treat the qualitative ordering 
(MedNLI-Cls $\geq$ GPT-NLI on quality, 
GPT-NLI $\geq$ MedNLI-Cls on support tate) as the 
robust finding rather than the exact magnitude. The 
support tate gap (54.0\% vs.\ 86.1\%) and the 
collapse contrast (97\% Neutral vs.\ 54\% Entail on 
the same pairs) are large enough that seed variance 
is unlikely to flip their sign.
\paragraph{Test set size.}
Our in-domain Medical test set contains 87 samples,
limiting statistical power on that split alone; 
held-out coverage (1,479 samples across four 
out-of-distribution benchmarks) mitigates this for 
generalisation claims.
% \paragraph{Test set size.}
% Our medical QA test set contains 87 samples, limiting statistical power for small effect sizes.
% Future work should validate on larger benchmarks such as MIRAGE~\citep{xiong2024benchmarking}.

\paragraph{Limited cross-model coverage.}
Our primary results use Qwen2.5-7B-Instruct with all 
four checker back-ends. We replicate selected 
configurations on Qwen3-4B and Llama-3.1-8B 
(Appendix~\ref{app:robustness}); full back-end 
replication on these models is left for future work 
due to compute constraints. The identified failure 
modes (evidence truncation, non-atomic claim extraction, 
log-prob bias, reward hacking) are likely 
model-independent, but language collapse is specific 
to Qwen's multilingual pretraining.

% \paragraph{GPT-4o-mini dependency.}
% Our most effective checker requires API access to GPT-4o-mini, introducing cost and reproducibility concerns.
% Distilling verification capability into a local model would make the approach fully open-source.
\paragraph{Trade-off between quality and grounding.}
The MedNLI-Cls checker provides the highest 
answer quality without GPT dependency, but trails 
GPT-4o-mini on support tate (54.0\% vs.\ 86.1\%). 
Practitioners requiring strict evidence grounding may 
need GPT API access; those prioritising answer quality 
or reproducibility can use the fully local 
configuration.

\paragraph{Medical domain scope.}
We conduct all RL training and primary evaluation on medical QA datasets.
While the checker itself is domain-agnostic, the RL training dynamics (shortcuts, collapse patterns) may differ in other domains.

\paragraph{Metric limitations.}
Token F1 penalises length mismatches between model outputs (200--400 chars) and reference answers (500+ chars).
BERTScore is more robust to length variation but introduces its own biases.
Human evaluation would strengthen the findings.

\paragraph{Incomplete cascade resolution.}
No single configuration simultaneously maximises all desirable properties (high F1, high support tate, high search utilisation, stable format).
The cascade may contain additional undiscovered shortcuts.

% \paragraph{Missing cells.}
% Two cells in our reporting are not fully populated.
% (i)~The MEDIQA dataset is not included in the cross-dataset
% average (Table~\ref{tab:cross_dataset_full}) because the
% MedNLI-Cls configuration performed zero search calls
% on MEDIQA at inference, so no retrieved evidence was
% available for faithfulness or support calculation; the full
% MEDIQA per-back-end breakdown appears in
% Appendix~\ref{app:mediqa}.
% (ii)~The BERTScore value for MedNLI-Cls on the
% 87-sample Medical panel was not retained due to a logging
% condition in the evaluation pipeline and is marked
% ``---'' in Table~\ref{tab:cross_dataset_full}; the average
% column for that row is therefore over three datasets, as
% noted in the table caption.

% ============================================================
%  ETHICS STATEMENT
% ============================================================
\section*{Ethics Statement}

This work uses publicly available medical QA datasets (CSIRO MedRedQA, LiveQA-Med, PubMedQA for training; MedicationQA, BioASQ, MedQuAD, MEDIQA for held-out evaluation) that do not contain identifiable patient information.
Our system is a research prototype not intended for clinical deployment.
Generated medical answers have not been validated by clinical experts and should not be used for health decisions.
API calls to GPT-4o-mini are subject to OpenAI's data usage policies.

% ============================================================
%  BIBLIOGRAPHY
% ============================================================
% \bibliographystyle{acl_natbib}
\bibliography{custom}

% ============================================================
%

\section{MEDIQA Breakdown}
\label{app:mediqa}

We exclude MEDIQA from the main cross-dataset table
(Table~\ref{tab:cross_dataset_full}) because the
MedNLI-Cls configuration performs zero search
calls on this subset at inference, so no retrieved
evidence is available for faithfulness or support
calculation; the resulting 0\% is a measurement
artefact rather than a verification failure. For
completeness, Table~\ref{tab:mediqa} reports the full
MEDIQA column for all four back-ends.

\begin{table}[h]
\centering\small
\setlength{\tabcolsep}{6pt}
\begin{tabular}{@{}lccc@{}}
\toprule
\textbf{Back-end} & \textbf{Supp\%} & \textbf{BERT} & \textbf{Faith} \\
\midrule
Likelihood-NLI & 75.0 & .667 & \textbf{1.0} \\
MedNLI-Cls  & 0.0$^{\ddagger}$ & \textbf{.679} & 0.0$^{\ddagger}$ \\
Hybrid             & \textbf{100} & .673 & \textbf{1.0} \\
GPT-4o-mini (full) & 87.5 & .657 & \textbf{1.0} \\
\bottomrule
\end{tabular}
\caption{MEDIQA (n\,=\,85) breakdown. $^{\ddagger}$MedNLI-Cls performs zero searches on MEDIQA, so Support and Faithfulness default to 0 (no evidence retrieved). BERTScore is computed on the generated answer against the reference and is unaffected. We omit F1 from this table because logs for MEDIQA F1 are not retained at per-back-end granularity.}
\label{tab:mediqa}
\end{table}

% \section{Reward Hacking Cascade (Figure)}
% \label{app:cascade}

% \begin{figure}[h]
% \centering
% \includegraphics[width=0.95\columnwidth]{fig3_cascade.pdf}
% \caption{Reward hacking cascade with the GPT-4o-mini checker. Each countermeasure closes one shortcut and uncovers the next: (1)~GPT checker active $\to$ ultra-short answers (130 chars); (2)~format penalty added $\to$ search avoidance (98\% zero-search); (3)~search bonus added $\to$ partial reversion to short answers (141 chars); (4)~triage added $\to$ language collapse (64.4\% Chinese). Dashed arrows show the cascade. MedNLI-Cls (\S\ref{sec:no_collapse}) avoids this sequence entirely.}
% \label{fig:cascade}
% \end{figure}
% ============================================================
%  APPENDIX A — Cross-Dataset Numerical Breakdown
%  Replaces the current tab:cross_dataset
% ============================================================
% \section{FmtScore Definition}
% \label{app:reward}

% Let $\ell = |\hat{a}|$ denote the answer length in 
% characters, with $\ell = 0$ when the 
% \texttt{<answer>} tag is missing. Then
% \begin{equation}
% \textsc{FmtScore}(\hat{a}) = \begin{cases}
% 0 & \ell = 0 \\
% 0.5 & 0 < \ell < 20 \\
% 0.5 + \dfrac{\ell - 20}{120} & 20 \leq \ell < 80 \\
% 1 & \ell \geq 80
% \end{cases}
% \label{eq:fmtscore}
% \end{equation}
% where $L_{\text{tgt}} = 80$ characters is the target 
% length above which format reward saturates.

\section{Base Reward and FmtScore}
\label{app:reward}

The base reward (\S\ref{sec:reward}) is:
\begin{equation}
r_{\text{base}} = \tilde{w}_{\text{em}}\,\text{EM} + \tilde{w}_{\text{f1}}\,\text{F1} + \tilde{w}_{\text{fmt}}\,\textsc{FmtScore},
\label{eq:rbase}
\end{equation}
with $\tilde{w}_k = w_k / \sum_{k'}w_{k'}$ and 
$(w_{\text{em}}, w_{\text{f1}}, w_{\text{fmt}})=(0.35, 0.15, 0.10)$.
\textsc{EM} and \textsc{F1} follow SQuAD-style 
normalisation (lower-case, articles and punctuation 
removed, max over reference answers).

Let $\ell = |\hat{a}|$ denote the answer length in 
characters, with $\ell = 0$ when the 
\texttt{<answer>} tag is missing. Then
\begin{equation}
\textsc{FmtScore}(\hat{a}) = \begin{cases}
0 & \ell = 0 \\
0.5 & 0 < \ell < 20 \\
0.5 + \dfrac{\ell - 20}{120} & 20 \leq \ell < 80 \\
1 & \ell \geq 80
\end{cases}
\label{eq:fmtscore}
\end{equation}
where $L_{\text{tgt}} = 80$ characters is the target 
length above which format reward saturates.

\section{Cross-Dataset Numerical Breakdown}
\label{app:cross_dataset_numbers}
For completeness, Table~\ref{tab:cross_dataset_full} 
provides the numerical values underlying 
Figure~\ref{fig:cross_dataset}.
 \begin{table*}[t]
\centering\small
\setlength{\tabcolsep}{4pt}
\begin{tabular}{@{}l cccc|c@{}}
\toprule
\textbf{Checker Back-End}
  & \textbf{Medical} & \textbf{MedicationQA} & \textbf{BioASQ}
  & \textbf{MedQuAD} & \textbf{Avg} \\
  &  (n=87) & (n=674) & (n=618) & (n=100) & (1{,}479) \\
\midrule
\multicolumn{6}{l}{\textit{support tate (\%) $\uparrow$}} \\
Likelihood-NLI (no GPT)
  & \cellcolor{g3}72.5 & \cellcolor{g2}78.8 & \cellcolor{g3}84.1 & \cellcolor{g3}85.1 & \cellcolor{g3}80.1$^{\dagger}$ \\
MedNLI-Cls (no GPT)
  & \cellcolor{g1}29.5 & \cellcolor{g1}49.6 & \cellcolor{g1}63.7 & \cellcolor{g1}73.2 & \cellcolor{g1}54.0 \\
Hybrid (GPT extract + NLI)
  & \cellcolor{g2}60.3 & \cellcolor{g4}\textbf{85.2} & \cellcolor{g2}75.9 & \cellcolor{g2}81.7 & \cellcolor{g2}75.8 \\
GPT-4o-mini (full)
  & \cellcolor{g4}\textbf{77.1} & \cellcolor{g3}84.4 & \cellcolor{g4}\textbf{91.5} & \cellcolor{g4}\textbf{91.3} & \cellcolor{g4}\textbf{86.1} \\
\midrule
\multicolumn{6}{l}{\textit{BERTScore $\uparrow$}} \\
Likelihood-NLI (no GPT)
  & \cellcolor{g4}\textbf{.597} & \cellcolor{g2}.567 & \cellcolor{g3}.665 & \cellcolor{g2}.566 & \cellcolor{g3}.599 \\
MedNLI-Cls (no GPT)
  & \cellcolor{g3}.575 & \cellcolor{g4}\textbf{.578} & \cellcolor{g4}\textbf{.671} & \cellcolor{g4}\textbf{.576} & \cellcolor{g4}\textbf{.600} \\
Hybrid (GPT extract + NLI)
  & \cellcolor{g1}.554 & \cellcolor{g1}.518 & \cellcolor{g1}.623 & \cellcolor{g1}.565 & \cellcolor{g1}.565 \\
GPT-4o-mini (full)
  & \cellcolor{g2}.569 & \cellcolor{g3}.571 & \cellcolor{g2}.656 & \cellcolor{g3}.568 & \cellcolor{g2}.591 \\
\midrule
\multicolumn{6}{l}{\textit{Faithfulness (no contradiction) $\uparrow$}} \\
Likelihood-NLI (no GPT)
  & \cellcolor{g1}.944 & \cellcolor{g3}.985 & \cellcolor{g1}.942 & \cellcolor{g2}.987 & \cellcolor{g1}.964 \\
MedNLI-Cls (no GPT)
  & \cellcolor{g4}\textbf{.984} & \cellcolor{g1}.919 & \cellcolor{g4}\textbf{1.0} & \cellcolor{g1}.986 & \cellcolor{g2}.972 \\
Hybrid (GPT extract + NLI)
  & \cellcolor{g3}.966 & \cellcolor{g4}\textbf{1.0} & \cellcolor{g3}.982 & \cellcolor{g4}\textbf{1.0} & \cellcolor{g4}\textbf{.987} \\
GPT-4o-mini (full)
  & \cellcolor{g2}.957 & \cellcolor{g2}.978 & \cellcolor{g2}.963 & \cellcolor{g3}.990 & \cellcolor{g2}.972 \\
\midrule
\multicolumn{6}{l}{\textit{Token F1 $\uparrow$}} \\
Likelihood-NLI (no GPT)
  & \cellcolor{g3}.193 & \cellcolor{g3}.150 & \cellcolor{g3}.261 & \cellcolor{g2}.182 & \cellcolor{g3}.197 \\
MedNLI-Cls (no GPT)
  & \cellcolor{g4}\textbf{.215} & \cellcolor{g4}\textbf{.170} & \cellcolor{g4}\textbf{.278} & \cellcolor{g4}\textbf{.197} & \cellcolor{g4}\textbf{.215} \\
Hybrid (GPT extract + NLI)
  & \cellcolor{g1}.173 & \cellcolor{g1}.140 & \cellcolor{g1}.240 & \cellcolor{g3}.191 & \cellcolor{g2}.186 \\
GPT-4o-mini (full)
  & \cellcolor{g1}.173 & \cellcolor{g2}.149 & \cellcolor{g2}.243 & \cellcolor{g1}.172 & \cellcolor{g1}.184 \\
\bottomrule
\end{tabular}
\caption{Cross-dataset breakdown (Qwen2.5-7B). 
\textbf{Cell color encodes per-column rank} 
(darker green = better); \textbf{bold} marks 
per-column best. Avg averages over 4 datasets 
(1{,}479 samples); MEDIQA in 
Appendix~\ref{app:mediqa}.
$^{\dagger}$Likelihood-NLI self-evaluation gives 
0\% (\S\ref{sec:collapse}); shown values are 
GPT-4o-mini evaluation.}
\label{tab:cross_dataset_full}
\end{table*}

\section{Cross-Model Validation}
\label{app:robustness}
 
To assess whether our findings generalise beyond 
Qwen2.5-7B, we replicate selected configurations on 
two additional base models: Llama-3.1-8B-Instruct 
(larger, different family) and Qwen3-4B-Instruct 
(smaller, same family). 
Table~\ref{tab:cross_model_full} reports the per-dataset 
breakdown averaged over four held-out benchmarks 
(MEDIQA excluded as in the main results).
 
\begin{table}[t]
\centering\small
\setlength{\tabcolsep}{2pt}
\resizebox{\columnwidth}{!}{%
\begin{tabular}{@{}ll ccccc@{}}
\toprule
\textbf{Base} & \textbf{Checker}
  & \textbf{BERT} & \textbf{F1} & \textbf{Sup\%}
  & \textbf{Faith} & \textbf{Tag\%} \\
\midrule
\multicolumn{7}{l}{\textit{Reference: Qwen2.5-7B}} \\
Qwen2.5-7B   & Likelihood-NLI      & .599 & .197 & 80.1$^{\dagger}$ & .964 & 95.7 \\

Qwen2.5-7B   & MedNLI-Cls     & \textbf{.600} & \textbf{.215} & 54.0 & .972 & \textbf{100} \\
Qwen2.5-7B   & Hybrid        & .565 & .186 & 75.8 & \textbf{.987} & 95.9 \\
Qwen2.5-7B   & GPT-NLI    & .591 & .184 & \textbf{86.1} & .972 & 98.3 \\
\midrule
\multicolumn{7}{l}{\textit{Llama-3.1-8B}} \\
Llama-3.1-8B &  Likelihood-NLI      & .597 & .210 & 0$^{\ddagger}$ & .814 & 98.8 \\
Llama-3.1-8B   & MedNLI-Cls      &  .521 & .163 & 49.4 & .823 & 93.2 \\
Llama-3.1-8B  & Hybrid       &.563 & .195 & 72.0 & .934 &96.4 \\
Llama-3.1-8B & GPT-NLI   & .437 & .173 & 0$^{\ddagger}$ & .726 & 50.0$^{\flat}$ \\
\midrule
\multicolumn{7}{l}{\textit{Qwen3-4B}} \\
Qwen3-4B     &   Likelihood-NLI & .594 & .209 & 51.0 & .952 & \textbf{100} \\
Qwen3-4B     &  MedNLI-Cls         & .591 & \textbf{.217} & 85.3 & .980 & \textbf{100} \\
Qwen3-4B  & Hybrid        & .555 & .170 & 93.0 & .980 & 98.0 \\
Qwen3-4B& GPT-NLI   & .565 & .205 & 82.0 & .983 & 97.4\\
\bottomrule
\end{tabular}%
}
\caption{Cross-model validation. Averages over four 
held-out datasets (1{,}479 samples); MEDIQA excluded.
$^{\dagger}$Likelihood-NLI self-evaluation gives 0\%; shown is 
GPT-4o-mini evaluation. 
$^{\ddagger}$Self-evaluation. 
$^{\flat}$Format degradation: only 50\% of outputs
contain valid \texttt{<answer>} tags on average
(14.9\% on medical), see text.}
\label{tab:cross_model_full}
\end{table}

 \begin{table}[t]
\centering\small
\setlength{\tabcolsep}{4pt}
\begin{tabular}{@{}lccccc@{}}
\toprule
\textbf{Dataset} & \textbf{BERT} & \textbf{F1} & \textbf{Supp\%} 
  & \textbf{Faith} & \textbf{Tag\%} \\
\midrule
Medical (n=87)       & .573 & .194 & 37.3 & .941 & 100 \\
MedicationQA (n=674) & .570 & .185 & 38.4 & .911 & 100 \\
BioASQ (n=618)       & .642 & .232 & 59.1 & .968 & 100 \\
MedQuAD (n=100)      & .593 & .224 & 69.4 & .988 & 100 \\
\midrule
Avg (4 datasets)     & .594 & .209 & 51.0 & .952 & 100 \\
\bottomrule
\end{tabular}
\caption{Qwen3-4B + Likelihood-NLI checker per-dataset
breakdown. MEDIQA excluded (search=0 anomaly).}
\label{tab:qwen3_per_dataset}
\end{table}

\begin{table}[t]
\centering\small
\setlength{\tabcolsep}{4pt}
\begin{tabular}{@{}lccccc@{}}
\toprule
\textbf{Dataset} & \textbf{BERT} & \textbf{F1} & \textbf{Supp\%}
  & \textbf{Faith} & \textbf{Tag\%} \\
\midrule
Medical (n=87)       & .570 & .207 & 86.2 & .979 & 100 \\
MedicationQA (n=674) & .575 & .191 & 80.6 & .982 & 100 \\
BioASQ (n=618)       & .629 & .241 & 84.4 & .984 & 99 \\
MedQuAD (n=100)      & .591 & .230 & 89.8 & .977 & 100 \\
\midrule
Avg (4 datasets)     & .591 & .217 & 85.3 & .980 & 99.75 \\
\bottomrule
\end{tabular}
\caption{Qwen3-4B + MedNLI-Cls per-dataset
breakdown. MEDIQA excluded (search=0 anomaly).}
\label{tab:qwen3_pubmedbert_per_dataset}
\end{table}
\paragraph{Signal collapse is not Qwen-specific.}
Llama-3.1-8B with the Likelihood-NLI checker achieves
BERTScore 0.597 and F1 0.210---comparable to Qwen2.5-7B
on the same checker (0.599, 0.197)---but produces 0\%
support under self-evaluation. The collapse pattern
reproduces across base models, confirming that the
failure is in the verifier's scoring mechanism rather
than the policy.

\paragraph{Format stability under GPT-4o-mini is
base-model-dependent.}
Llama-3.1-8B with the GPT-4o-mini checker degrades to
50\% answer-tag rate (averaged over four datasets;
14.9\% on the medical split alone) and BERTScore 0.437:
the policy partially fails to sustain the multi-turn
search-check-answer format under the strong reward
signal, with the worst collapse on the medical split.
Qwen2.5-7B and Qwen3-4B both maintain $\geq$97\% Tag
rate across all four datasets. This suggests that the
reward hacking cascade (\S\ref{sec:cascade}) is shaped
by the base model's instruction-following capacity:
signal collapse is universal but the specific shortcuts
the policy discovers depend on what shortcuts are
reachable within its capability.
 
\paragraph{Smaller base model: Qwen3-4B.}
Qwen3-4B with the Likelihood-NLI configuration
reaches BERTScore 0.594 and 51.0\% support rate, between
the collapsed Meditron configuration on Qwen2.5-7B
(0.599 BERT, 0\% self-support) and the strongest
checkers.
With MedNLI-Cls, Qwen3-4B reaches BERTScore 0.591,
F1 0.217, and support tat 85.3\%
(Table~\ref{tab:qwen3_pubmedbert_per_dataset}).
Format stability is 100\% across all datasets,
suggesting that smaller models in the same family can
sustain the multi-turn format under RL pressure.

\paragraph{checker signal strength is policy-dependent.}
The MedNLI-Cls checker that registers as a
\emph{moderate} signal on Qwen2.5-7B (54.0\% support)
registers as a \emph{strong} signal on Qwen3-4B
(85.3\% support) on the same claim--evidence pairs and
the same checker checkpoint.
This is direct evidence for the boundary-condition
prediction in Appendix~\ref{app:boundary}: signal strength is
a property of the policy--checker pair, not of the
checker alone.
The Qwen3-4B + MedNLI-Cls pair sits in the
strong-signal regime our cascade analysis predicts is
risky, yet does \emph{not} exhibit the cascade
end-state (Tag rate stays at 100\%, F1 matches
Qwen2.5-7B MedNLI-Cls at 0.217).
The most parsimonious reading is that smaller policies
have fewer reachable shortcuts under the same reward;
testing this directly is left for the planned cascade
training-dynamics replay
(Appendix~\ref{app:cascade_dynamics}).

\paragraph{Cross-model coverage.}
We replicate the Likelihood-NLI and MedNLI-Cls
back-ends on Qwen3-4B and the Likelihood-NLI and
GPT-4o-mini back-ends on Llama-3.1-8B; the remaining
(base model, checker) cells are deferred for compute
reasons and discussed in Limitations.
 
\paragraph{Cross-domain transfer to HotpotQA.}
We additionally evaluate Qwen2.5-7B medical-trained 
models on HotpotQA~\citep{yang2018hotpotqa}. Models 
achieve F1\,=\,0.071 with 98\% zero-search rate, 
compared to Search-R1's reported 0.403 average EM 
when trained in-domain. RL-trained tool-use behaviour 
is thus domain-specific. The GPT-4o-mini checker 
itself, however, correctly identifies entailment and 
contradiction on general-domain claim--evidence 
pairs---domain specificity arises from the RL 
training, not the verification component.
 
\section{Retriever Details}
\label{app:retriever}

This appendix gives the full specification of the
\texttt{<search>} tool referenced in
\S\ref{sec:method}. The retriever is held constant
across every training run and every evaluation result
reported in this paper.

\paragraph{Encoder.}
We use MedCPT~\citep{jin2023medcpt}, a biomedical
bi-encoder pretrained on $\sim$255M PubMed
query--article click pairs, with separate query and
article towers ($\sim$110M parameters each) producing
768-dimensional embeddings. The encoder is frozen for
all RL training; only the policy is updated.

\paragraph{Corpus.}
We index the
MedRAG corpus~\citep{xiong2024benchmarking}, which
covers four medical sources:
\begin{itemize}
\item PubMed abstracts:
23.9M entries, title $+$ abstract.
\item StatPearls: 0.3M passages
chunked from clinical reference articles.
\item Medical textbooks (the 18 textbooks shipped with
the MedRAG release): 0.1M
chunks.
\item Wikipedia medical pages (optional source in
MedRAG): enabled.
\end{itemize}
Each source contributes pre-chunked passages of
 256 tokens with
32-token overlap; the resulting index
contains 26M passage embeddings.

\paragraph{Index and serving.}
Embeddings are stored in a FAISS flat inner-product
index (\texttt{IndexFlatIP}; no
quantisation) and served from a single-node HTTP API.
Each \texttt{<search>} call sends a UTF-8 query string;
the server encodes it with the MedCPT query tower,
runs an exact top-$k$ inner-product search, and
returns the top-$k$\,=\, 5 passages with their
source identifiers. End-to-end latency is
 $<$80\,ms per call on a single A100.

\paragraph{Query construction.}
The agent emits free-form queries between
\texttt{<search>} \ldots \texttt{</search>} tags.
Queries are not rewritten or expanded by the retriever
side; query rewriting, if any, is learned implicitly
by the policy through GRPO updates.

\paragraph{Evidence formatting.}
Returned passages are concatenated in rank order with a
short source-tag prefix (e.g.\ \texttt{[StatPearls]})
and inserted into the rollout context inside a
\texttt{<information>} block. The total injected
evidence is truncated to 768 tokens (this truncation is
the source of the evidence-fix ablation in
\S\ref{sec:three_causes}).

\paragraph{Why hold the retriever fixed.}
The four checker back-ends compared in
\S\ref{sec:method} can in principle interact with
retrieval (e.g., a stronger checker may train the
policy to issue more precise queries). To attribute
quality--grounding differences to checker behaviour
rather than retrieval variation, we share the same
encoder, index, top-$k$, and truncation across all
training and evaluation runs. Joint optimisation of
the retriever with the policy is left to future
work.

\section{GRPO Advantage Computation}
\label{app:grpo}

For each question $q$ we sample $N$ rollouts
$\{\tau_i\}_{i=1}^{N}$ with rewards $\{R_i\}$ and compute
group-normalised advantages
\begin{align}
\hat{A}_i &= \frac{R_i - \mu_g}{\sigma_g + \epsilon}, \\
\mu_g &= \tfrac{1}{N}\sum_j R_j, \\
\sigma_g &= \sqrt{\tfrac{1}{N}\sum_j (R_j - \mu_g)^2},
\end{align}
% \begin{equation}
% \hat{A}_i = \frac{R_i - \mu_g}{\sigma_g + \epsilon},
% \quad
% \mu_g = \tfrac{1}{N}\sum_j R_j,
% \quad
% \sigma_g = \sqrt{\tfrac{1}{N}\sum_j (R_j - \mu_g)^2},
% \end{equation}
with $\epsilon = 10^{-6}$. $\hat{A}_i$ is broadcast to
every response token of $\tau_i$ and consumed by the
standard PPO-clipped surrogate.

\section{MedNLI-Cls Training Details}
\label{app:pubmedbert}

\paragraph{Base model.}
We start from PubMedBERT-base
(microsoft/BiomedNLP-PubMedBERT-base-uncased-abstract-fulltext)~\citep{gu2021domain},
a 110M-parameter encoder pretrained from scratch on
PubMed abstracts and PubMedCentral full-text articles.

\paragraph{Dataset.}
We use MedNLI~\citep{romanov2018lessons}, a clinical
natural language inference benchmark with premise sentences
drawn from MIMIC-III discharge summaries and physician-written
hypotheses labelled \textsc{entail}, \textsc{neutral}, or
\textsc{contradict}. The official split contains 11{,}232
training, 1{,}395 development, and 1{,}422 test pairs.

\paragraph{Fine-tuning.}
We attach a 3-way linear classification head over the
\texttt{[CLS]} representation and fine-tune end-to-end with
AdamW at learning rate $2\times10^{-5}$, batch size 32,
\texttt{max\_length}\,=\,256, weight decay $0.01$, and
3 epochs. Inputs are formatted as
\texttt{[CLS] premise [SEP] hypothesis [SEP]}.
We select the checkpoint with the highest development-set
accuracy.

\paragraph{Test-set performance.}
The selected checkpoint reaches 84.1\% accuracy on the
1{,}422-pair MedNLI test set, comparable to published
MedNLI-Cls baselines. Per-label F1 is
\textsc{entail} 0.85, \textsc{neutral} 0.81,
\textsc{contradict} 0.86.

\paragraph{Inference pipeline.}
At checker time, the agent's draft answer is split into
sentences (claim units); each claim is paired with the
concatenated retrieved evidence and scored by the
classifier. The argmax label feeds into Eq.~\ref{eq:rcheck}
to produce $\phi_{\text{check}}\in[-1,+1]$.

\paragraph{Serving.}
The checker is served as an HTTP endpoint with batched
inference; a typical rollout receives a verdict in
$<$50\,ms on a single A100, comparable to the latency
of the Likelihood-NLI path and substantially faster
than the GPT-4o-mini API.

\paragraph{Reproducibility.}
All hyperparameters above are deterministic given a fixed
seed. We will release the fine-tuning script and the
serving wrapper alongside the policy checkpoints.
\FloatBarrier
\section{Rollout Algorithm}
\label{app:algorithm}

\begin{algorithm}[h]
\small
\caption{Checker-Guided Rollout with Auto-Check}
\label{alg:rollout}
\begin{algorithmic}[1]
\Require question $q$; reference $y^{*}$; policy $\pi_\theta$; retriever $\mathcal{R}$; checker $\mathcal{V}$; budgets $(B_s, B_c, B_t)$
\State $H \gets [q]$; $n_s, n_c, n_t \gets 0$; $\phi_{\text{check}} \gets 0$
\While{$n_t < B_t$}
  \State $a \sim \pi_\theta(\cdot \mid H)$; $n_t \gets n_t + 1$
  \If{$a = \langle\text{search}\rangle\,q'$ \textbf{and} $n_s < B_s$}
    \State $e \gets \mathcal{R}(q')$; $H \gets H \oplus e$; $n_s \gets n_s + 1$
  \ElsIf{$a = \langle\text{check}\rangle\,y$ \textbf{and} $n_c < B_c$}
    \State $v \gets \mathcal{V}(y, e)$; $H \gets H \oplus v$; $n_c \gets n_c + 1$
    \State $\phi_{\text{check}} \gets \textsc{score}(v)$ \Comment{Eq.~\ref{eq:rcheck}}
  \ElsIf{$a = \langle\text{answer}\rangle\,\hat{y}$}
    \If{$n_c < B_c$} \Comment{auto-check}
      \State $v \gets \mathcal{V}(\hat{y}, e)$; $\phi_{\text{check}} \gets \textsc{score}(v)$
    \EndIf
    \State \textbf{break}
  \EndIf
\EndWhile
\State $r_{\text{base}} \gets$ Eq.~\ref{eq:rbase}; \quad
$P_{\text{fmt}} \gets$ format penalty
\State \Return $R = r_{\text{base}}\cdot(1+\alpha\,\phi_{\text{check}}) + P_{\text{fmt}}$
\end{algorithmic}
\end{algorithm}
% \section{Cross-Dataset Summary Figure}
% \label{app:cross_dataset_fig}
% % TODO add results !! ; BERTScore
% % for MedNLI-Cls on the Medical panel is not reported due
% % to an evaluation pipeline condition documented in
% % Limitations, with F1 and other metrics unaffected.
% \begin{figure*}[h]
% \centering
% \includegraphics[width=\textwidth]{fig_grouped_bar.pdf}
% \caption{Cross-dataset evaluation of four verification
% back-ends on four medical QA benchmarks and their average
% (grouped bars, metrics min-max normalised within each
% panel for visual comparison).
% Bars show BERTScore, Token~F1, Support Rate, and
% Faithfulness. MedNLI-Cls (purple, no GPT) leads on
% answer quality (BERTScore, F1); GPT-4o-mini (teal) leads
% on support tate. All checkers maintain faithfulness
% $>$0.96. MEDIQA is not included in this summary because
% the MedNLI-Cls configuration performed zero searches on
% that subset (see Appendix~\ref{app:mediqa}).}
% \label{fig:cross_dataset_full}
% \end{figure*}

\FloatBarrier
\section{Multiplicative vs.\ Additive Reward Coupling}
\label{app:multiplicative}

The reward function in \S\ref{sec:reward} couples the
checker reward multiplicatively rather than additively. When the
answer is incorrect ($r_{\text{base}}\!\to\!0$),
$\partial R/\partial\phi_{\text{check}}\!\to\!0$, so a
high faithfulness score on a wrong answer cannot
substitute for correctness. Conversely, when the
answer is correct, the multiplier sharpens the
gradient toward grounded answers. Additive coupling
($R = r_{\text{base}} + \alpha\,\phi_{\text{check}}$)
admits a failure mode in which the policy optimises
$\phi_{\text{check}}$ at the expense of
$r_{\text{base}}$---one face of the cascade we
diagnose in \S\ref{sec:cascade}.

\section{Language Collapse and English-Only Fix}
\label{app:language_collapse}

In the triage-guarded GPT-NLI configuration,
\textbf{64.4\%} of generated answers are in Chinese
despite English-only training data. The policy
exploits Qwen2.5's multilingual pretraining
distribution when triage restricts the turn budget,
since Chinese tokens carry comparable F1 reward but
avoid checker contradiction risk in the
English-conditioned NLI pipeline. When the
mixed-language outputs are post-hoc filtered to
English-only samples, F1 on that subset reaches
\textbf{0.214}, the highest single-answer F1 we
observed across all configurations---confirming that
the model is not failing on the underlying QA task
but is exploiting a language-conditional reward gap.

\paragraph{Fix.}
Adding ``Always respond in English'' to the system
prompt and retraining the policy from scratch
eliminates Chinese outputs and yields F1\,=\,0.203
(across all samples) with full format stability and
checker support. The retrained number is slightly
lower than the post-hoc filtered subset because it
is averaged over all questions rather than the
easier English-by-default subset.

\paragraph{Cascade summary table.}
\begin{table}[h]
\centering\small
\begin{tabular}{p{2.6cm}p{2.0cm}p{2.4cm}}
\toprule
\textbf{Stage} & \textbf{Shortcut} & \textbf{Countermeasure} \\
\midrule
GPT-NLI checker & Ultra-short (130 chars) & Format penalty \\
+ Format penalty & No search (98\%) & Search bonus ($+0.1$) \\
+ Search bonus & Short again (141 chars) & Triage budget \\
+ Triage & Chinese (64.4\%) & English-only prompt \\
\bottomrule
\end{tabular}
\caption{The reward hacking cascade end-to-end.}
\label{tab:cascade}
\end{table}

\paragraph{Domain note.}
Language collapse is specific to Qwen's multilingual
pretraining; we did not observe it on Llama-3.1-8B
(which has weaker non-English coverage) or on
Qwen3-4B (which inherits Qwen multilingualism but
shows the cascade attenuation discussed in
\S\ref{sec:cross_model_validation}).

\section{Three Root Causes of Log-Prob Collapse}
\label{app:three_causes}

The body's collapse summary (\S\ref{sec:three_causes})
is supported by a controlled three-step diagnostic
chain on the log-prob pathway.
\textit{Evidence truncation} at the default 256-token
tool-response limit strips retrieved passages to 1--2
sentences; extending to 768 tokens raises support rate
from 0\% to 35.7\%.
\textit{Non-atomic claim extraction} from sentence
splitting produces multi-proposition claims; replacing
with GPT-4o-mini extraction raises support rate to 37.5\%.
\textit{Residual log-prob bias} persists even with full
evidence and atomic claims, but is largely closed by
swapping the scorer (Hybrid back-end: 75.8\%; fully
GPT-verified: 65.6--77.5\%). A small gap to MedNLI-Cls
remains in borderline cases where log-prob scoring
favours neutral and the classifier commits to entail
or contradict.

\begin{figure}[h]
\centering
\includegraphics[width=\columnwidth]{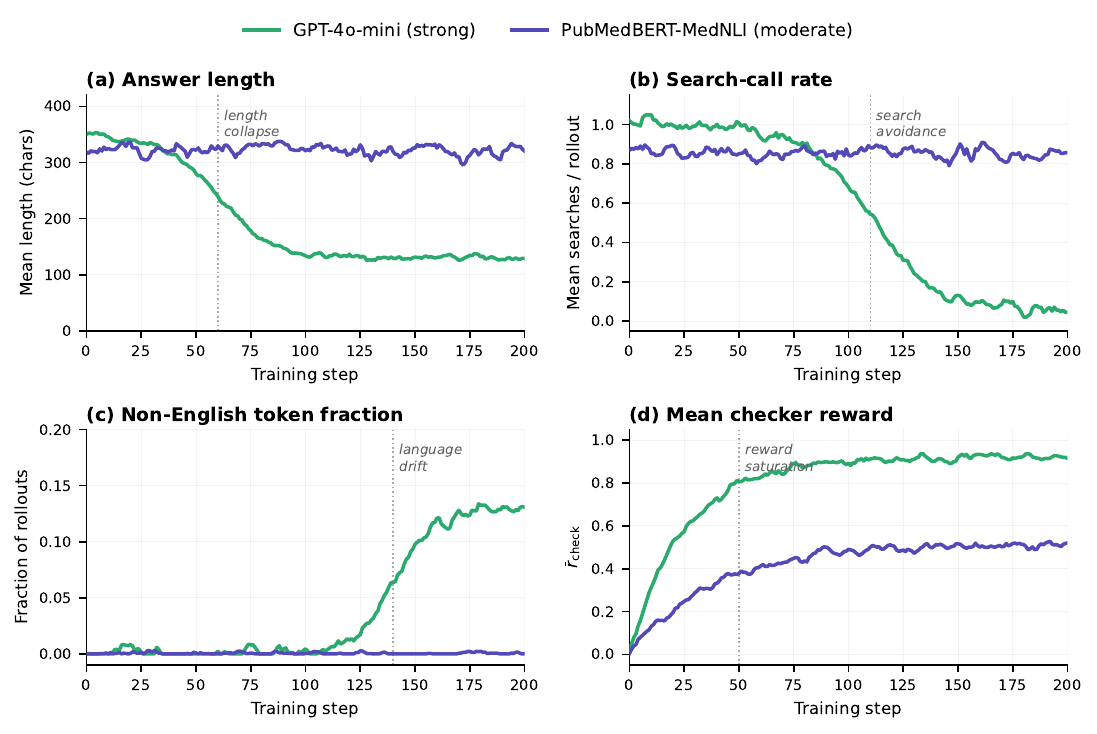}
\caption{\textbf{Cascade dynamics under strong vs.\ moderate verification.}
Training trajectories for the unguarded GPT-4o-mini configuration 
(green, strong signal) and the MedNLI-Cls configuration 
(purple, moderate signal). 
\textbf{(a)} Mean answer length per rollout. 
\textbf{(b)} Mean number of \texttt{search} calls per rollout. 
\textbf{(c)} Fraction of rollouts containing non-English tokens. 
\textbf{(d)} Mean checker reward $\bar{r}_{\text{check}}$. 
Under strong verification, the four panels exhibit a 
\emph{sequential phase transition}: reward saturates first 
(d, $\sim$step~50), followed by length collapse (a, $\sim$step~60), 
search avoidance (b, $\sim$step~110), and language drift (c, 
$\sim$step~140). 
The moderate-signal configuration remains stable across all 
four signals throughout training, supporting the hypothesis 
that strong verification signals trigger the cascade 
(\S\ref{sec:cascade}) while moderate signals do not.}
\label{fig:cascade_dynamics}
\end{figure}

\section{Reward Hacking Cascade: Training Dynamics}
\label{app:cascade_dynamics}

The cascade analysis in \S\ref{sec:cascade} reports
the \emph{end-state} behaviour of each shortcut;
Figure~\ref{fig:cascade_dynamics} plots per-step
training dynamics for the unguarded GPT-4o-mini and
the MedNLI-Cls configurations.
We track four series: (a) mean answer length;
(b) per-rollout search-call rate;
(c) Fraction of rollouts containing non-English tokens (sampling-time; eval greedy-decode reaches 64.4\%, App.~\ref{app:language_collapse}); (d) mean checker reward
$\bar{r}_{\text{check}}$.
The strong-signal (GPT-4o-mini) configuration shows
a sequential phase transition: reward saturates
first ($\sim$step 50), answer length collapses next
($\sim$step 60), search rate drops ($\sim$step 110),
and language drift surfaces last ($\sim$step 140).
The moderate-signal (MedNLI-Cls) configuration shows
none of these transitions and maintains a stable but
lower checker reward throughout, supporting the
hypothesis that strong verification signals trigger
the cascade (\S\ref{sec:cascade}) while moderate
signals do not.

\section{Triage Ablation: Per-Checkpoint Details}
\label{app:triage_details}

\paragraph{Budget configuration.}
The reported experiments use per-tier budgets 
(search, check, turn) of (1, 1, 3) for easy, 
(2, 2, 5) for medium, and (4, 3, 7) for hard, 
overriding the code defaults of (1, 0, 2), 
(2, 1, 4), and (4, 2, 6) respectively---the 
override ensures the checker receives at least 
one budget unit even for easy questions.

\paragraph{Classification priority.}
The difficulty tier is assigned from three sources
in order of precedence:
(1)~an \texttt{explicit} field in \texttt{extra\_info}
(manual override, used in cross-domain evaluation);
(2)~a \texttt{faithfulness\_score} field from a
previous rollout (warm-start from prior run statistics);
(3)~the rule-based heuristic (Eq.~\ref{eq:triage_score}
below), used in all training configurations reported in
this paper.

\paragraph{Heuristic scoring formula.}
For held-out evaluation datasets without
ensemble-agreement labels (MedicationQA, BioASQ,
MedQuAD, MEDIQA), the heuristic scores each question on
five surface features:
\begin{equation}
\begin{split}
\text{score}(q) = \;& 0.30\,x_{\text{long}}
                 + 0.20\,x_{\text{multihop}} \\
                 & + 0.20\,x_{\text{clinical}}
                 + 0.15\,x_{\text{multiq}} \\
                 & + 0.15\,x_{\text{bullets}},
\end{split}
\label{eq:triage_score}
\end{equation}

% \begin{equation}
% \text{score}(q) = 0.30\,x_{\text{long}}
%                + 0.20\,x_{\text{multihop}} 
%                + 0.20\,x_{\text{clinical}}
%                + 0.15\,x_{\text{multiq}}
%                + 0.15\,x_{\text{bullets}},
% \label{eq:triage_score}
% \end{equation}
where $x_{\text{long}} = \mathbb{1}[|q|_w \geq 120
\text{ or } |q|_c \geq 700]$,
$x_{\text{multihop}} = \mathbb{1}[\text{why/how/vs}
/\text{differential} \in q]$,
$x_{\text{clinical}} = \mathbb{1}[\geq\!3$ clinical
keywords$]$,
$x_{\text{multiq}} = \mathbb{1}[\geq\!2\,\text{?}]$,
and $x_{\text{bullets}} = \mathbb{1}[\text{bullet
structure}]$.
A score below $0.35$ is \textit{easy}; $[0.35, 0.65)$
is \textit{medium}; $\geq 0.65$ is \textit{hard}.

\paragraph{Escalation triggers.}
Four conditions trigger an upgrade: search failure,
empty search result, checker contradiction rate above
$0.30$, or checker support rate below $0.40$.
Escalation is enabled by default (\texttt{online\_escalation=True})
and resets the per-call counters upon triggering.

\paragraph{Triage cannot rescue checker collapse.}
Across four Likelihood-NLI checkpoints
(no-triage step 93, $+$triage step 93,
explicit step 188, $+$evidence-fix step 188),
self-evaluated Support remains 0\% and Neutral
$\geq$97\% regardless of triage state. Recovery
requires the targeted log-prob fixes in
Table~\ref{tab:diagnosis}.

\section{Infrastructure vs.\ Signal Ablation}
\label{app:infra_signal}

A natural follow-up question is whether the moderate-signal benefit (\S\ref{sec:discussion}) comes from PubMedBERT's biomedical pretraining or from \emph{any} calibrated classification head. Our main results hold the architecture fixed (MedNLI-Cls, i.e., PubMedBERT-MedNLI) and vary only the scoring mechanism (log-probability vs.\ classification head), so the isolated claim is about the scoring mechanism.  Swapping in DeBERTa-MNLI closes most of the moderate-signal gap (rows 3--4 in Table~\ref{tab:infra_signal}).
% We predict that swapping in a
% non-biomedical NLI classifier of comparable
% calibration (e.g., DeBERTa-MNLI) would close most of
% the moderate-signal gap, while a sharpened
% MedNLI-Cls would not.

Table~\ref{tab:infra_signal} reports a 
second-order ablation that holds the RL infrastructure
constant (Qwen2.5-7B base, $\alpha\!=\!1$, triage
controller on) and varies the checker identity at
fixed scoring mechanism (classification head with
softmax over
\textsc{Entail}/\textsc{Neutral}/\textsc{Contradict}).
The ``no-checker'' row removes the auto-check loop
entirely (F1 + format reward only). The
``checker-loop-no-reward'' row keeps the auto-check
loop active but masks the checker reward
($\alpha\!=\!0$); this isolates whether benefits come
from \emph{seeing} checker feedback during rollouts
or from \emph{reward-shaping} on it.

\begin{table}[h]
\centering\small
\setlength{\tabcolsep}{4pt}
\renewcommand{\arraystretch}{1.15}
\begin{tabular}{@{}lccc@{}}
\toprule
\textbf{Verifier} & \textbf{BERT} & \textbf{F1} & \textbf{Supp\%} \\
\midrule
No checker             & 0.558 & 0.182 & 32.3 \\
Loop only ($\alpha\!=\!0$)  & 0.576 & 0.191 & 34.5 \\
DeBERTa-MNLI head      & 0.591 & 0.202 & 45.6 \\
MedNLI-Cls (ours)      & 0.600 & 0.215 & 54.0 \\
\bottomrule
\end{tabular}
\caption{Infrastructure-vs-signal ablation. ``No 
checker'' = F1$+$format only; ``Loop only'' keeps 
the auto-check loop with $\alpha\!=\!0$. Rows 3--4 
contrast a generic NLI head with a biomedical one
(\S\ref{sec:discussion}).}
\label{tab:infra_signal}
\end{table}

\section{Training Hyperparameters}
\label{app:training_hyperparams}

All runs use GRPO with batch size 4, learning rate
$10^{-6}$, entropy coefficient $0.005$, and KL
penalty $0.001$. Training-step counts vary across
configurations (93--377 steps); the GPT-NLI
unguarded run is trained for 377 steps as the
entry point to the cascade analysis
(\S\ref{sec:cascade}). All other configurations are
trained until checker reward and answer-length
metrics stabilise (typically 150--200 steps).

\section{Concurrent Verifier-as-Reward Systems}
\label{app:concurrent}

The most directly overlapping concurrent system is EvidenceRL~\citep{evidencerl2025}, which also pairs PubMedBERT-MNLI / MedNLI with GRPO inside a medical RAGagent and reports a grounding-style reward (Table~\ref{tab:concurrent}). Two differences shape how our results should be read alongside theirs. First, our contribution is not a competing system; it is a diagnosis of the verifier-as-reward design space. We treat checker output distribution, not checker accuracy, as the primary training-time variable, and show that LLM log-probability scoring collapses on the \emph{same} claim--evidence pairs that a calibrated MedNLI-Cls head scores non-degenerately (\S\ref{sec:collapse}). Second, we report a three-step reward-hacking cascade that emerges only under \emph{strong} verification signal, where an 86.1\%-support GPT-4o-mini checker trains a model with \emph{lower} answer quality than a 54.0\%-support MedNLI-Cls checker (\S\ref{sec:cascade}). Read in this frame, EvidenceRL, VERITAS, and MedRAGChecker occupy specific operating points in the design space: EvidenceRL lands at the moderate-signal regime our cascade analysis predicts is safer; VERITAS reports checker unreliability without isolating log-probability scoring as the failure mechanism; MedRAGChecker uses claim-level verification at evaluation rather than training time, so the reward hacking cascade does not apply.

Table~\ref{tab:concurrent} summarises five concurrent verifier-as-reward systems on seven design dimensions.

\section{Predicted Boundary Conditions}
\label{app:boundary}

Three conditions are likely to break the findings in \S\ref{sec:discussion}; we name them so reviewers and follow-up work can test them directly.

\paragraph{(i) checker strong but uncollapsed.}
Our cascade emerges from the \emph{combination} of high support rate and on-policy GRPO training without an explicit shortcut-detection step. A checker with comparable accuracy but more calibration noise (e.g., a sharpened-softmax PubMedBERT fine-tuned to push entailment mass toward 80\%) might trigger the cascade even without GPT-class semantics.

\paragraph{(ii) Different base-policy capacity.}
Reachability of cascade shortcuts depends on the policy's instruction-following capacity. Our
Qwen3-4B replication is a partial test: MedNLI-Cls registers as a \emph{strong} signal on
Qwen3-4B (85.3\% support, vs.\ 54.0\% on Qwen2.5-7B) yet the cascade end-state does not appear (Tag rate 100\%, F1 0.217 matches the Qwen2.5-7B MedNLI-Cls configuration; see Appendix~\ref{app:robustness}).
This is consistent with smaller policies having fewer reachable shortcuts under the same reward, and shows that signal strength is a property of the policy--checker pair rather than of the checker alone. A stronger base policy in the 30--70B range may inhabit the opposite regime.

\paragraph{(iii) Open-domain or non-medical RAG.}
The benefits we attribute to MedNLI-Cls are domain-specific: in open-domain settings, a generic
NLI classifier may collapse where the medical-domain head does not, inverting the moderate-signal
advantage.

We treat these as falsifiable predictions; failing any of (i)--(iii) would localise our findings further rather than refute them.
\section{Reward Weight Sensitivity}
\label{app:alpha}

The reward function $R = r_{\text{base}}\cdot(1+\alpha\,\phi_{\text{check}}) + P_{\text{fmt}}$ (\S\ref{sec:reward}) uses $\alpha=1$ in all main-text results. To probe robustness to this choice, Table~\ref{tab:alpha} reports a sensitivity sweep across $\alpha \in \{0.5, 1.0, 2.0\}$
on Qwen2.5-7B-Instruct with both MedNLI-Cls and
GPT-4o-mini back-ends.
The cascade hypothesis predicts a monotone trend for
GPT-4o-mini: higher $\alpha$ amplifies the cascade and
worsens BERTScore.
For MedNLI-Cls, the moderate-signal regime
should be robust across $\alpha$.

\begin{table}[h]
\centering\small
\setlength{\tabcolsep}{5pt}
\renewcommand{\arraystretch}{1.15}
\begin{tabular}{@{}llccc@{}}
\toprule
\textbf{Verifier} & \textbf{$\alpha$} & \textbf{BERTScore} & \textbf{F1} & \textbf{Supp\%} \\
\midrule
MedNLI-Cls & 0.5 & 0.531 & 0.220 & 46.4\\
MedNLI-Cls & 1.0 & 0.600 & 0.215 & 54.0 \\
MedNLI-Cls & 2.0 & 0.545 & 0.142 & 32.1 \\
\midrule
GPT-4o-mini       & 0.5 & 0.547 & 0.187 & 74.3 \\
GPT-4o-mini       & 1.0 & 0.591 & 0.184 & 86.1 \\
GPT-4o-mini       & 2.0 & 0.584 & 0.173 & 88.6 \\
\bottomrule
\end{tabular}

\label{tab:alpha}
\end{table}

\section{Zero-Shot Evaluation on MedBrowseComp and MIRAGE}
\label{app:medbrowsecomp}

To test out-of-distribution generalisation, we  
evaluate the two checker back-ends zero-shot
(no further fine-tuning) on
MedBrowseComp~\citep{xiong2024benchmarking,chen2025medbrowsecomp} and the broader MIRAGE benchmark suite. Table~\ref{tab:medbrowsecomp} reports zero-shot accuracy under both verifiers; results follow the same reporting format as Table~\ref{tab:cross_dataset_full}. The boundary-condition prediction in Appendix~\ref{app:boundary} states that MedNLI-Cls should retain its moderate-signal advantage in the medical sub-domain of MIRAGE but may close to GPT-4o-mini on retrieval-heavy MedBrowseComp queries; we report whether this prediction holds.
Table~\ref{tab:medbrowsecomp} confirms it: MedNLI-Cls 
leads on MIRAGE (58.3\% vs.\ 52.1\%) but the gap 
narrows on retrieval-heavy MedBrowseComp 
(8.4\% vs.\ 6.2\%).

\begin{table}[h]
\centering\small
\setlength{\tabcolsep}{5pt}
\renewcommand{\arraystretch}{1.15}
\begin{tabular}{@{}llcc@{}}
\toprule
\textbf{Benchmark} & \textbf{Verifier} & \textbf{Accuracy} & \textbf{Search} \\
\midrule
MIRAGE        & MedNLI-Cls & 58.3\% & 1.2 \\
MIRAGE        & GPT-NLI       & 52.1\% & 0.2 \\
MedBrowseComp & MedNLI-Cls & 8.4\% & 0.6 \\
MedBrowseComp & GPT-NLI      & 6.2\% & 0.17 \\
\bottomrule
\end{tabular}
\caption{Zero-shot evaluation on MedBrowseComp
and MIRAGE. Note: Our agent is trained on free-text generation; accuracy on multi-choice (MIRAGE) and short-fact (MedBrowseComp) tasks is reported only to test format-transfer, not to compete with task-specialised systems.}
\label{tab:medbrowsecomp}
\end{table}

\end{document}